\documentclass[journal,twocolumn]{IEEEtran}

\usepackage{graphicx,times,amsmath,amssymb,cite,subfigure,stfloats,booktabs, url,multirow,afterpage}

\usepackage{graphicx}
\usepackage{caption}

\usepackage{array}
\usepackage{booktabs}
\usepackage{multirow}
\usepackage{array}
\usepackage{makecell}
\usepackage{subfigure} 

\usepackage{commath,amsfonts,algorithm,algorithmic}
\usepackage{cite,multirow,bm,multicol,booktabs,threeparttable,extpfeil}

\allowdisplaybreaks

\begin{document}

\title{MIRepNet: A Pipeline and Foundation Model\\ for EEG-Based Motor Imagery Classification}

\author{%
  Dingkun~Liu\textsuperscript{\dag}, 
  Zhu~Chen\textsuperscript{\dag}, 
  Jingwei Luo, 
  Shijie Lian, 
  Dongrui~Wu\textsuperscript{*},~\IEEEmembership{Fellow,~IEEE}%
  \thanks{\textsuperscript{\dag}These authors contributed equally to this work.}%
  \thanks{\textsuperscript{*}Corresponding author: Dongrui Wu (drwu09@gmail.com).}%
  \thanks{D.~Liu, Z.~Chen, J.~Luo and D.~Wu are with the Ministry of Education Key Laboratory of Image Processing and Intelligent Control, School of Artificial Intelligence and Automation, Huazhong University of Science and Technology, Wuhan 430074, China.}%
\thanks{S.~Lian is with the School of Computer Science and Technology, Huazhong University of Science and Technology, Wuhan 430074, China.}
  \thanks{D.~Liu, S.~Lian and D.~Wu are also with Zhongguancun Academy, Beijing 100094, China.}%
}

\maketitle

\begin{abstract}
Brain-computer interfaces (BCIs) enable direct communication between the brain and external devices. Recent EEG foundation models aim to learn generalized representations across diverse BCI paradigms. However, these approaches overlook fundamental paradigm-specific neurophysiological distinctions, limiting their generalization ability. Importantly, in practical BCI deployments, the specific paradigm such as motor imagery (MI) for stroke rehabilitation or assistive robotics, is generally determined prior to data acquisition. This paper proposes MIRepNet, the first EEG foundation model tailored for the MI paradigm. MIRepNet comprises a high-quality EEG preprocessing pipeline incorporating a neurophysiologically-informed channel template, adaptable to EEG headsets with arbitrary electrode configurations. Furthermore, we introduce a hybrid pretraining strategy that combines self-supervised masked token reconstruction and supervised MI classification, facilitating rapid adaptation and accurate decoding on novel downstream MI tasks with fewer than 30 trials per class. Extensive evaluations across five public MI datasets demonstrated that MIRepNet consistently achieved state-of-the-art performance, significantly outperforming both specialized and generalized EEG models. Our code will be available on GitHub\footnote{https://github.com/staraink/MIRepNet}.
\end{abstract}


\section{Introduction}

\vspace{3.5pt}
\begingroup
\setlength{\parindent}{0pt}
\itshape\fontsize{10}{12}\selectfont
\begin{center}
``Nature, to be commanded, must be obeyed."\\[3pt] 
\end{center}
\normalfont\raggedleft\scshape --- Francis Bacon\par
\endgroup
\vspace{4pt}

A brain-computer interface (BCI) establishes a direct communication pathway between the brain and external devices \cite{nicolas2012brain}. Electroencephalogram (EEG) is one of the most widely used non-invasive modalities for BCIs due to its cost-effectiveness, ease of use, and safety \cite{kannathal2005characterization,vourvopoulos2016usability,drwuPIEEE2023}. However, the practical deployment of EEG-based BCIs faces significant challenges, including substantial individual differences, limited availability of high-quality training data, and discrepancies in electrode configurations across different EEG headsets \cite{drwuMITLBCI2022, liu2025spatial}. Consequently, traditional approaches typically require extensive calibration data collection from each user, posing a substantial obstacle to real-world usability.

\begin{figure}[htbp]
\centering
\begin{minipage}[b]{0.5\linewidth}
  \centering
  \raisebox{0.4mm}{\includegraphics[width=\linewidth]{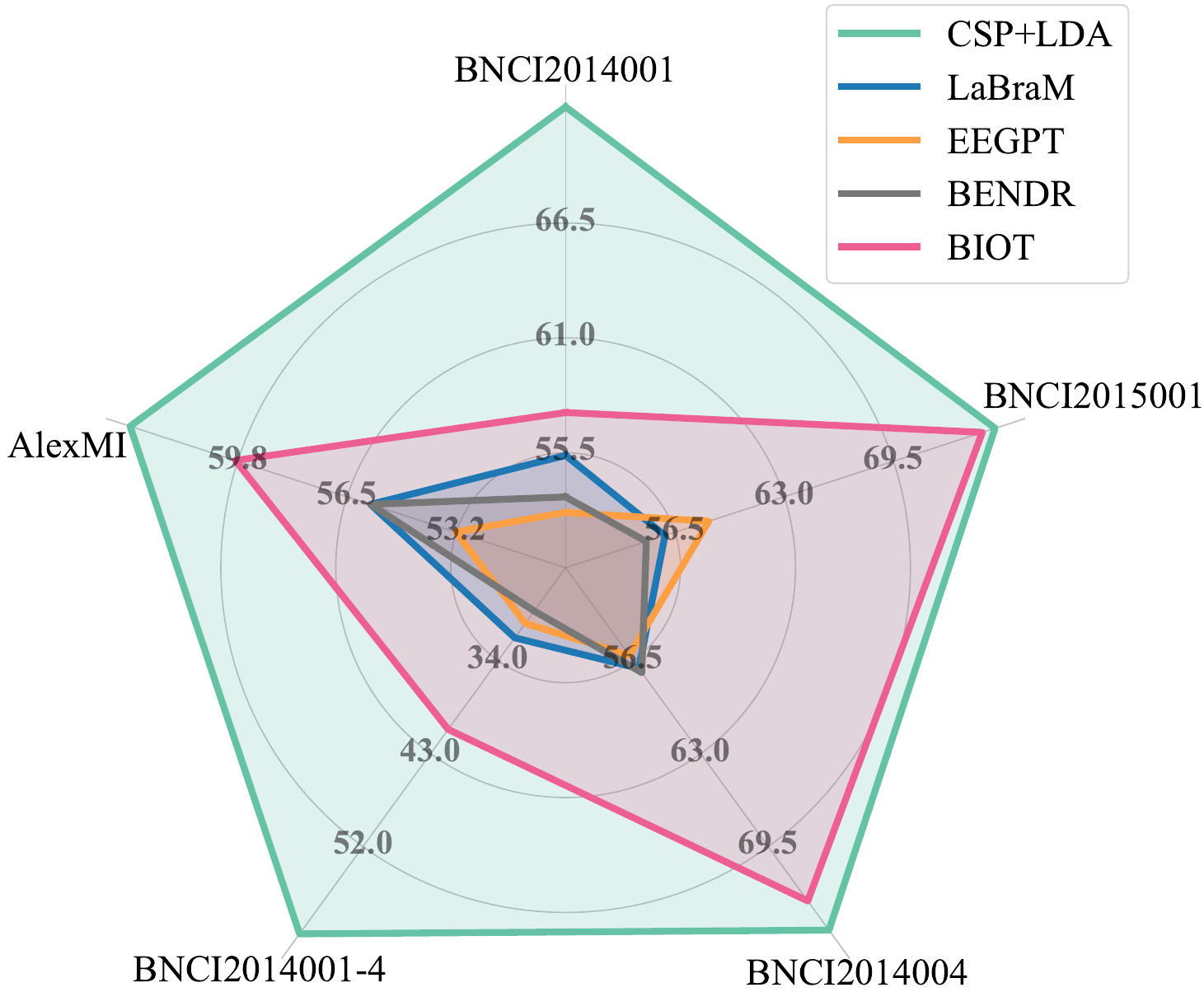}}\\[-2pt]
  \makebox[\linewidth][l]{\hspace*{0.43\linewidth}(a)}
\end{minipage}\hfill
\begin{minipage}[b]{0.5\linewidth}
  \centering
  \raisebox{0.4mm}{\includegraphics[width=\linewidth]{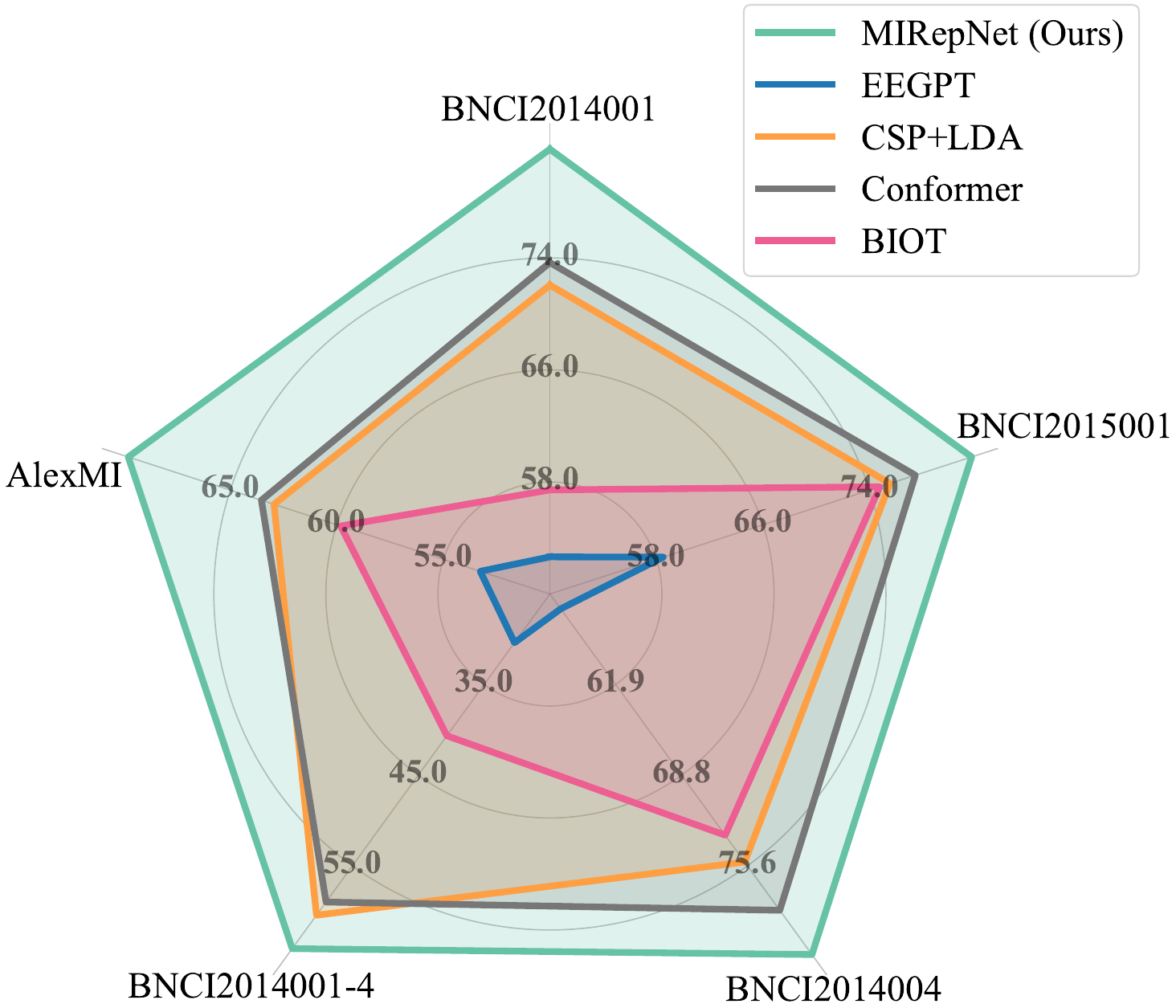}}\\[-2pt]
  \makebox[\linewidth][l]{\hspace*{0.43\linewidth}(b)}
\end{minipage}
\caption{Visualization of results on five MI downstream datasets after 30\% finetuning for each subject. Specialist models were trained from scratch. (a) Comparison of CSP+LDA with EEG generalist models; (b) Comparison of MIRepNet (ours) with competitive EEG specialist and generalist baselines.}
\label{fig:radar}
\end{figure}

Recently, generalized EEG foundation models \cite{wan2023eegformer, yang2023biot, wang2024cbramod, jiang2024large, wang2025eegpt} have emerged as promising solutions for addressing calibration challenges by learning universal EEG representations from large-scale datasets. Such foundation models aim to facilitate effective adaptation across diverse downstream BCI tasks. However, existing EEG generalized models face several inherent limitations:

\begin{enumerate}
\item Existing models typically merge EEG signals from various paradigms, such as motor imagery (MI), steady-state visual evoked potentials (SSVEP), and event-related potentials (ERP), during pretraining. Nevertheless, distinct BCI paradigms inherently involve fundamentally different neurophysiological mechanisms, characterized by paradigm-specific active cortical regions and frequency bands \cite{liu2025clean}. Combining multiple paradigms into a single pretraining stage contradicts the intrinsic neurophysiological organization of EEG signals, resulting in suboptimal representations and limited downstream generalization performance.

\item In practice, the paradigm associated with a downstream task is typically predetermined before data acquisition. For instance, stroke patients requiring exoskeleton assisted rehabilitation naturally align with an MI-based system \cite{li2024survey}, whereas epilepsy monitoring necessitates a dedicated epilepsy-specific paradigm and corresponding model \cite{hermann2017paradigm}. Consequently, when the user profile encompassing patient population and intended task is accessible, selecting a paradigm-specific foundation model for direct adaptation becomes both feasible and preferable.

\item More critically, current generalized models require extensive post-training involving other subjects' data from the entire downstream dataset (usually comprising 8–12 subjects) to adapt effectively to a new subject. Compared to a scenario in which only 30\% of a single new subject's data is needed for finetuning, the adaptation cost in terms of data volume increases dramatically (at least 20 times). Such extensive post-training may even underperform conventional subject-specific training on the same data (see Fig.~\ref{fig:radar} (a)). Ideally, a foundation model should adapt directly to any new subject or task without additional large-scale post-training.
\end{enumerate}

These insights motivate the development of paradigm-specific foundation models that support efficient adaptation using limited data from a new user. Such models promise significantly enhanced decoding accuracy and substantially reduced calibration requirements. Among existing BCI paradigms, MI is particularly noteworthy due to its extensive exploration and broad clinical applications, including stroke rehabilitation and assistive technologies such as smart wheelchairs \cite{mokienko2014motor}.

\begin{figure}[t]
\centering
\includegraphics[width=0.9\columnwidth]{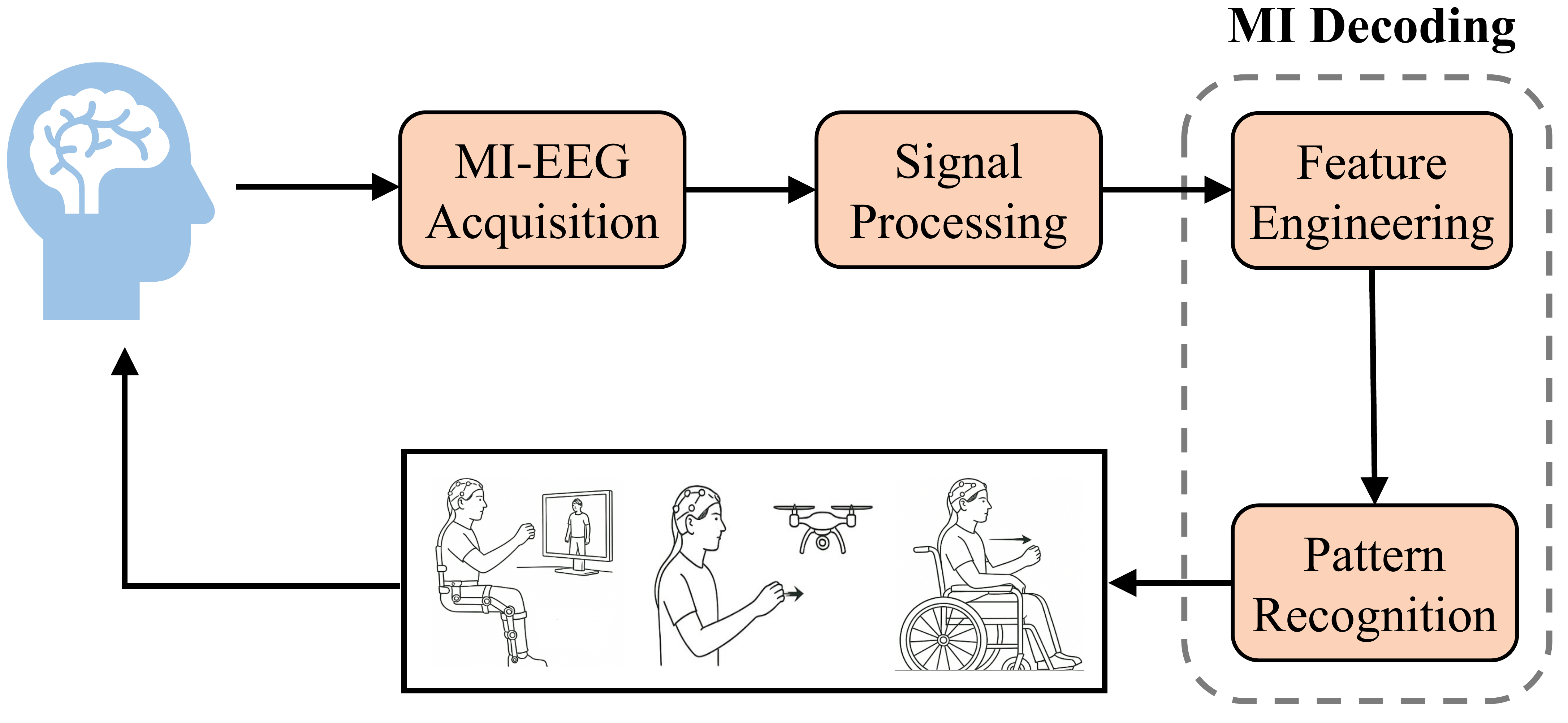}
\caption{A closed-loop MI-based BCI system.}
\label{closed-loop}
\end{figure}

\begin{figure*}[t]
\centering
\includegraphics[width=\textwidth]{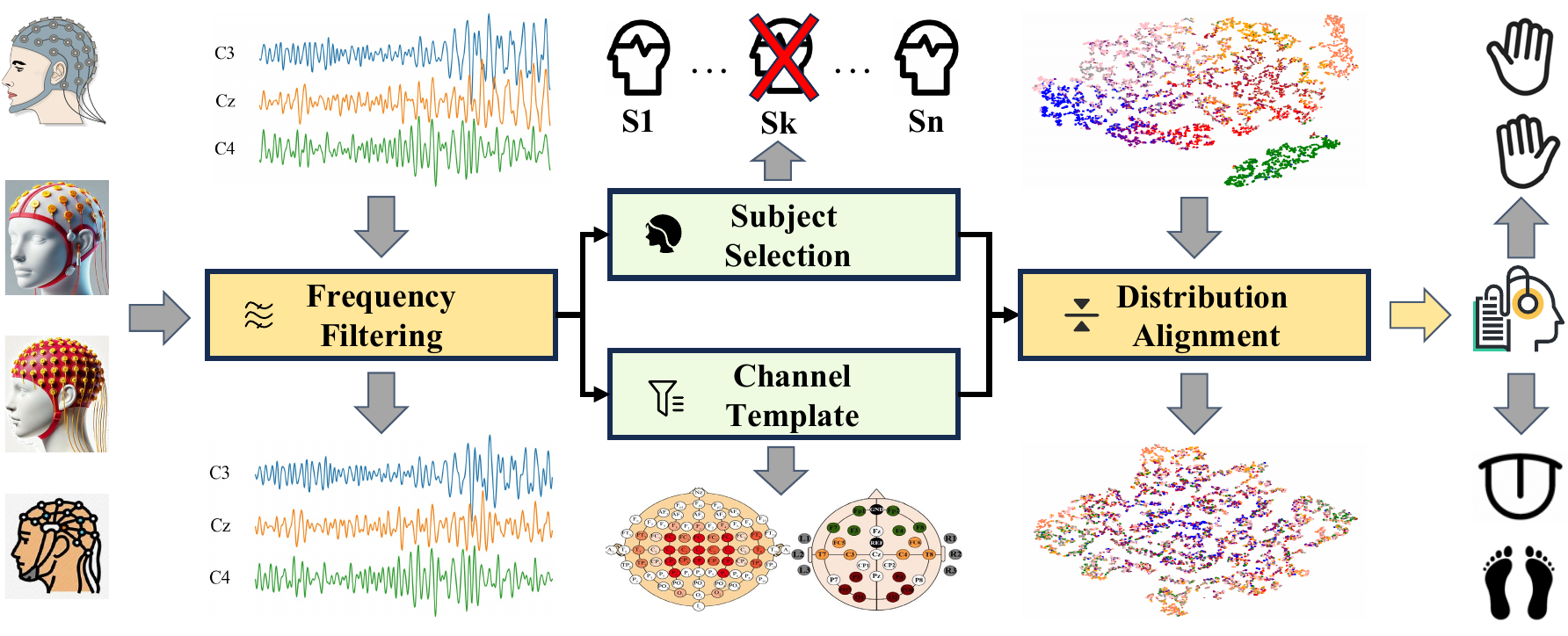}
\caption{High quality MI data construction pipeline.}
\label{CLEANMI}
\end{figure*}

This paper proposes \emph{MIRepNet}, an EEG foundation model explicitly designed for MI (whose pipeline is shown in Fig.~\ref{closed-loop}). MIRepNet leverages neurophysiological insights of the MI paradigm to construct a tailored preprocessing and representation learning pipeline, significantly enhancing the generalizability and efficiency of EEG decoding.

The main contributions of this work are:
\begin{enumerate}
\item We introduce MIRepNet, the first foundation model specifically tailored for MI tasks. By capturing MI-specific neurophysiological features, MIRepNet effectively learns generalizable representations for MI decoding.
\item We propose a high-quality EEG preprocessing pipeline comprising subject screening, a unified channel-template-based spatial alignment, frequency filtering, temporal resampling, and distribution alignment. This approach addresses challenges arising from heterogeneous EEG headset configurations, ensuring data consistency across diverse datasets.
\item We develop an efficient pretraining approach combining masked token reconstruction and supervised MI classification. This strategy enables the model to acquire robust, generalizable temporal-spatial EEG representations.
\end{enumerate}
Extensive experiments on five public MI datasets including 47 downstream subjects demonstrated that MIRepNet achieved state-of-the-art decoding accuracy (see Fig.~\ref{fig:radar} (b)). Moreover, MIRepNet required significantly fewer calibration trials (fewer than 30 trials per class) and rapidly converged in a few epochs, highlighting its practical utility and effectiveness.
    
\section{Related Work}

\subsection{MI Decoding Algorithms}

Convolutional neural networks (CNNs) have achieved significant success in EEG decoding. EEGNet~\cite{lawhern2018eegnet} utilizes compact depthwise separable convolutions to efficiently extract discriminative EEG features. ShallowConvNet and DeepConvNet~\cite{schirrmeister2017deep} integrate temporal and spatial convolutions with nonlinear transformations to capture log-band-power features. FBCNet~\cite{mane2021fbcnet} employs filter-bank spectral filtering combined with depthwise spatial convolutions, effectively modeling spectro-spatial-temporal patterns. IFNet~\cite{wang2023ifnet} explicitly captures cross-frequency interactions via interactive frequency convolution modules, while EEG-Conformer~\cite{song2022eeg} merges local convolutional features with global self-attention mechanisms. However, these specialized methods typically require extensive labeled data from individual subjects or specific acquisition setups, limiting their generalizability and practical deployment.

\subsection{EEG Generalized Models}

Recent EEG foundation models seek to achieve robust transferability across diverse downstream tasks. BIOT \cite{yang2023biot} tokenizes multi-channel biosignals into unified ``sentences" and employs Transformer-based cross-dataset pretraining. BENDR \cite{kostas2021bendr} utilizes a wav2vec-style self-supervised Transformer for contrastive representation learning directly from large-scale raw EEG signals. CBraMod~\cite{wang2024cbramod} employs a criss-cross transformer with masked EEG reconstruction. Similarly, LaBraM \cite{jiang2024large} adopts semantic EEG tokenization and masked modeling to capture generalizable EEG representations. EEGPT \cite{wang2025eegpt} leverages spatio-temporal alignment and masked reconstruction tasks to learn universal EEG embeddings, while NeuroLM \cite{jiangneurolm} frames EEG representation learning within a language-modeling paradigm.

However, these mixed-paradigm foundation models often fail to learn truly paradigm-agnostic representations due to fundamental neurophysiological distinctions among different BCI paradigms. Moreover, adapting such generalized models to novel downstream tasks typically necessitates additional post-training involving the entire downstream dataset (including all subjects), significantly increasing data requirements, elevating privacy concerns, and adversely affecting user-friendliness.

\begin{figure*}[t]
\centering
\includegraphics[width=\textwidth]{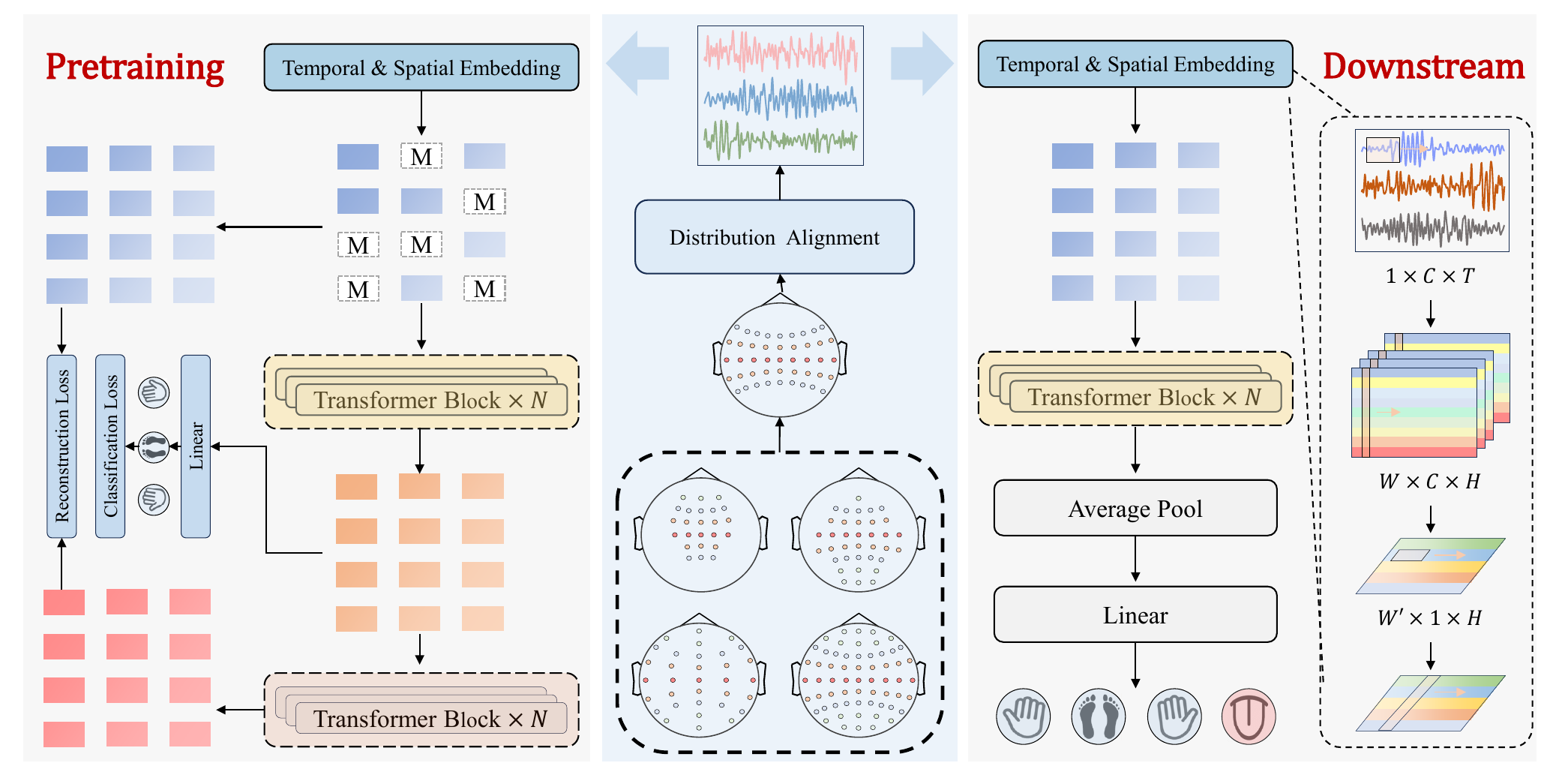}
\caption{Overview of MIRepNet. Heterogeneous MI-EEGs from different headsets are first unified to a common channel template and whitened via distribution alignment. The aligned trial is converted into temporal–spatial tokens. A proportion $\alpha$ of tokens are masked and reconstructed with a Transformer encoder–decoder, while the same encoder is simultaneously supervised by an MI classification head. The joint objective learns precise, generalizable MI representations that adapt with only few-shot data.}
\label{MIRepNet}
\end{figure*}

\section{Method}

We introduce MIRepNet, a foundation model specifically developed for MI paradigm. MIRepNet aims to learn robust and generalizable temporal-spatial EEG representations that facilitate rapid and accurate adaptation to novel MI downstream tasks. To achieve this, we construct a high-quality MI-specific EEG data pipeline, introduce a spatially-informed channel template, and propose a hybrid pretraining scheme combining self-supervised masked reconstruction and supervised MI classification.

\subsection{Problem Definition}

Given $r$ labeled pretraining datasets $\{\{(X_{i,j}^\mathrm{s}, y_{i,j}^\mathrm{s})\}_{i=1}^{n_\mathrm{s}^j}\}_{j=1}^{r}$, where $X_{i,j}^\mathrm{s}\in\mathbb{R}^{C_\mathrm{s}^j \times T_\mathrm{s}^j}$ and $y_{i,j}^\mathrm{s} \in \{1, 2, \ldots, \mathcal{C}\}$ ($\mathcal{C}$ is the number of categories), and $u$ downstream target datasets $\{\{X_{i,k}^\mathrm{t}\}_{i=1}^{n_\mathrm{t}^k}\}_{k=1}^{u}$, in which $X_{i,k}^\mathrm{t} \in \mathbb{R}^{C_\mathrm{t}^k \times T_\mathrm{t}^k}$, the goal is to pretrain a foundation model capable of rapidly adaptation to accurately predict the labels $\{y_{i,k}^\mathrm{t}\}$ of new trials in downstream MI datasets.

\subsection{High-Quality MI-EEG Data Construction}
\subsubsection{Temporal Filtering and Alignment}

MI decoding primarily leverages modulations in sensorimotor rhythms (SMRs), specifically within the $\alpha$ (8-13 Hz) frequency bands~\cite{neuper2006motor}. Therefore, we first apply a standard band-pass filter of 8–30 Hz to extract these task-relevant frequency components. Due to varying sampling frequencies across different datasets, we subsequently resample all EEG trials to a unified sampling rate $f_{\mathrm{target}}$, typically set to 250 Hz, for consistent temporal alignment across datasets:

\begin{multline}
    \label{eq:mne_resample}
    \bar{X}_{i,j}^\mathrm{s} = \mathrm{Resample}\left( X_{i,j}^\mathrm{s},\, f_{\mathrm{orig}}^j,\, f_{\mathrm{target}} \right) \in \mathbb{R}^{C_\mathrm{s}^j \times T}, \\
    \forall j \in \{1,\dots,r\}
\end{multline}

\subsubsection{Subject Selection}

EEG data quality typically exhibits significant variation among subjects due to differences in engagement, fatigue, and/or recording conditions. To maintain high-quality data for foundation model training, we implement a rigorous subject screening protocol. Specifically, for each subject, we train a preliminary within-subject classifier using only their own trials. Subjects whose accuracy falls below a predefined performance threshold are excluded from subsequent analysis. This procedure effectively eliminates data compromised by inattention, artifact contamination, or poor signal integrity, yielding a robust dataset with reliable and stable EEG characteristics.

\subsubsection{Channel Template}

EEG headset configurations differ substantially in terms of electrode placement and count, posing challenges for integrating heterogeneous datasets. Based on MI-specific neurophysiological knowledge, we define a standard channel template consisting of electrodes positioned over the frontal-central (FC), central (C), centro-parietal (CP), and temporal (T) regions, denoted by a fixed set of $C$ electrodes. For trials from datasets with different electrode configurations, we perform spatial interpolation using inverse-distance weighting.

Formally, for each EEG trial $X_{i,j}^\mathrm{s}\in\mathbb{R}^{C_s^j\times T}$ recorded from channels $\{e_k\}_{k=1}^{C_s^j}$, we first calculate Euclidean distances between the electrode coordinates:
\begin{equation}
d_{ik}=\lVert\phi(t_i)-\phi(e_k)\rVert_2,\,i=1,\dots,C,\,k=1,\dots,C_s^j,
\end{equation}
where $\phi(\cdot)$ maps electrode indices to the 2D scalp coordinate. 

We then compute the interpolation weights:
\begin{equation}
W_{ik} =
\begin{cases}
1, & \exists\,k^*: d_{ik^*}=0,\ k = k^*,\\[4pt]
0, & \exists\,k^*: d_{ik^*}=0,\ k \neq k^*,\\[6pt]
\dfrac{d_{ik}^{-1}}{\sum_{l=1}^{C_s^j}d_{il}^{-1}}, & \text{otherwise},
\end{cases}
\label{eq:weights}
\end{equation}
where $i=1,\dots,C$ and $k=1,\dots,C_s^j$. 

Finally, the spatially-aligned EEG trials become:
\begin{multline}
\label{eq:channel_interp}
X'[b,i,t]
=\sum_{c=1}^{C_s}W_{ij}\,\bar{X}[b,c,t],\\
b=1,\dots,B,\;i=1,\dots,C,\;t=1,\dots,T.
\end{multline}

This channel template approach standardizes input EEG signals, preserving MI-related spatial information and enabling unified modeling.

\subsubsection{Distribution Alignment}

EEG signals inherently exhibit non-stationarity and inter-subject variability. To reduce marginal distribution shifts, we adopt Euclidean alignment (EA) \cite{he2019transfer,drwuEA2025} via whitening transformations. For a given subject with $n$ processed trials $\{X'_i\in\mathbb{R}^{C\times T}\}_{i=1}^n$, we first compute the reference covariance matrix:
\begin{equation}
\bar{R} \;=\; \frac{1}{n}\sum_{i=1}^n X'_i (X'_i)^\top,
\label{eq:EA-Ref}
\end{equation}
and then normalize each trial by
\begin{equation}
\widetilde{X}_i \;=\; \bar{R}^{-1/2} \,X'_i,
\quad i=1,\dots,n.
\label{eq:EA-Whitening}
\end{equation}
After EA, $\frac{1}{n}\sum_{i=1}^n \widetilde{X}_i \widetilde{X}_i^\top = I$,
so the transformed EEG trials $\{\widetilde{X}_i\}$ from different subjects have the identity covariance, effectively mitigating shifts in second‐order statistics. 

In all subsequent steps, we replace the original trials $\{X_i\}$ with the aligned signals $\{\widetilde{X}_i\}$ to ensure a more consistent input distribution. The effectiveness of EA is demonstrated by $t$-SNE visualization in Fig.~\ref{CLEANMI}.

\begin{table*}[!t] \centering
\caption{Summary of the pretraining and downstream MI-EEG datasets.}
\label{tab:datasets}
\begin{tabular}{c|c|c|c|c|c|c}
\toprule
  &\multirow{2}{*}{Dataset} & Number of & Number of & Sampling & Number of  & \multirow{2}{*}{Categories} \\
  & & Subjects & Channels & Rate (Hz) & total trials &  \\
\midrule
\multirow{7}{*}{Pretraining}
&BNCI2014002 & 14 & 15 & 512 & 2240 &  right hand, both feet \\
&PhysionetMI & 109 & 64 & 160 & 7373 &  left hand, right hand, feet \\
&Dreyer2023 & 87 & 27 & 512 & 16152 &  left hand, right hand \\
&Weibo2014 & 10 & 60 & 200 & 2370 &  left hand, right hand, feet \\
&Zhou2016 & 4 & 14 & 250 & 900 &  left hand, right hand, feet \\
&Lee2019 & 54 & 62 & 1000 & 10800 &  left hand, right hand \\
&Cho2017 & 52 & 64 & 512 & 10520 &  left hand, right hand \\

\midrule
\multirow{5}{*}{Downstream}
&BNCI2014001 & 9 & 22 & 250  & 1296 &  left hand, right hand \\
&BNCI2015001 & 12 & 13 & 512 & 2400 &  right hand, both feet \\
&BNCI2014004 & 9 & 3 & 250 & 1400 &  left hand, right hand \\
&AlexMI & 8 & 16 & 512 & 320 & left hand, right hand \\
&BNCI2014001-4 & 9 & 22 & 250  & 2592 & left hand, right hand, feet, tongue \\
\bottomrule
\end{tabular}
\end{table*}

\subsection{Temporal–Spatial Representation}

The architecture of MIRepNet is shown in Fig.~\ref{MIRepNet}. Given an EEG trial $X\in\mathbb R^{C\times T}$, we first encode the temporal dynamics through a convolutional embedding $\phi_{t}$, with kernel size $k_{t}$ and stride $s_{t}$, yielding:
\begin{equation}
U \;=\;\phi_{t}(X)
\;=\;\mathrm{Conv}_{t}(X)
\;\in\;\mathbb R^{C\times H\times W},
\end{equation}
where $W=\lceil T/s_{t}\rceil$. We then rearrange $U$ to $U'\;\in\;\mathbb R^{W\times C\times H}$. A subsequent spatial convolutional encoder $\phi_{s}$ compresses the spatial information, resulting in:
\begin{equation}
S
\;=\;\phi_{s}(U')
\;=\;\mathrm{Conv}_{s}(U')
\;\in\;\mathbb R^{W'\times 1\times H},
\end{equation}
where \(W'\) depends on the spatial convolution parameters. Finally, we apply temporal average-pooling and a \(1\times1\) convolution to generate compact tokens:
\begin{equation}
Z
\;=\;\mathrm{Conv}_{1\times1}\bigl(\mathrm{AvgPool}(S)\bigr)
\;\in\;\mathbb R^{D\times 1\times H'},
\end{equation}
reshaped as a sequence of $H'$ tokens $\{\,z_{i}\in\mathbb R^{D}\}_{i=1}^{H'}$, each capturing integrated temporal-spatial EEG features.

\subsection{MIRepNet Pretraining}

\subsubsection{Masked Token Reconstruction}

After obtaining $H'$ temporal-spatio tokens $\{z_i\in\mathbb R^{D}\}_{i=1}^{H'}$ from each EEG trial, we apply a masked reconstruction task to encourage the model to learn robust contextual representations. 

Specifically, let $\alpha\in(0,1)$ be the mask ratio.  We uniformly sample a mask set $\mathcal{M}\subset\{1,\dots,H'\}$ with $|\mathcal{M}|=\alpha H'$, and define the masked token sequence
\begin{equation}
\tilde z_i =
\begin{cases}
\texttt{[MASK]}, & i\in\mathcal{M},\\
z_i,            & i\notin\mathcal{M},
\end{cases}
\end{equation}
where \(\texttt{[MASK]}\in\mathbb R^D\) is a learnable embedding.  The masked tokens $\{\tilde z_i\}$ are fed into an $M$-layer Transformer encoder $f_{\theta}$, producing contextual embeddings $\{c_i\}$.  A lightweight Transformer decoder $g_{\phi}$ then reconstructs the full token set:
\begin{equation}
\hat z_i \;=\; g_{\phi}\bigl(\{c_j\}_{j=1}^{H'}\bigr)\,,\quad i=1,\dots,H'.
\end{equation}

We then minimize the reconstruction loss:
\begin{equation}
\mathcal L_{\mathrm{rec}}
=\frac{1}{|\mathcal{M}|}
\sum_{i\in\mathcal{M}}
\bigl\lVert \hat z_i - z_i\bigr\rVert_2^2.
\end{equation}

By predicting the original tokens from their context, the model learns to capture coherent temporal-spatial patterns, which improves downstream MI decoding performance.

\subsubsection{Supervised MI Classification}

To further enhance the representation discriminability, we attach a supervised classification head $f_{\mathrm{cls}}\colon\mathbb R^D\to\mathbb R^C$ to the Transformer encoder. Given the pooled representation
\begin{equation}
v \;=\;\frac{1}{H'}\sum_{i=1}^{H'}c_i
\;\in\;\mathbb R^D,
\end{equation}
the head produces class logits $s=f_{\mathrm{cls}}(v)\in\mathbb R^C$.  For a true label $y\in\{1,\dots,C\}$, we define the cross‐entropy loss as:
\begin{equation}
\mathcal L_{\mathrm{cls}}
=-\log\frac{\exp\bigl(s_{y}\bigr)}{\sum_{k=1}^{C}\exp\bigl(s_{k}\bigr)}.
\end{equation}
During pretraining, $\mathcal L_{\mathrm{cls}}$ is averaged over all labeled trials and combined with the masked‐reconstruction loss to jointly optimize the model.

\subsubsection{Joint Pretraining}

MIRepNet is trained by jointly minimizing the masked token reconstruction loss $\mathcal{L}_{\mathrm{rec}}$ and the MI classification loss $\mathcal{L}_{\mathrm{cls}}$. Formally, the overall pretraining loss is
\begin{equation}
\mathcal{L}_{\mathrm{pretrain}}
= \mathcal{L}_{\mathrm{rec}} \;+\; \mathcal{L}_{\mathrm{cls}}.
\end{equation}

\begin{table*}[htpb]     \centering
\fontsize{9}{12}\selectfont %
       \caption{Accuracies (\%) in BNCI2014001. The best accuracies are marked in bold, and the second best by an underline.}  \label{tab:14001}
    \begin{tabular}{w{c}{3.2cm}|w{c}{2.1cm}|*{9}{w{c}{0.69cm}}|w{c}{1.3cm}}   \toprule
        Setting & Approach & S0 & S1 & S2 & S3 & S4 & S5 & S6 & S7 & S8 & Avg. \\
        \midrule
         \multirow{9}{*}{\makecell{EEG Specialist Models\\(30\% Trained)}}
         & CSP+LDA & 81.18 & 53.47 & 93.07 & 63.37 & 53.47 & \underline{66.34} & 66.34 & \textbf{96.04} & 75.25 & 72.06 \\
        ~ & ShallowConv & 75.91 & 52.81 & \underline{95.38} & 54.46 & 50.17 & 59.41 & 57.10 & 91.75 & 83.50 & 68.94$_{\pm1.15}$ \\
        ~ & DeepConv & 66.34 & 50.17 & 80.20 & 55.78 & 49.51 & 54.79 & 51.16 & 65.02 & 67.66 & 60.07$_{\pm1.65}$ \\
        ~ & EEGNet & 60.73 & 52.48 & 82.51 & 51.49 & 55.78 & 52.81 & 53.14 & 68.32 & 84.16 & 62.38$_{\pm2.19}$ \\
        ~ & IFNet & 80.86 & 51.49 & 92.08 & 57.10 & 55.45 & 56.11 & 58.42 & 90.10 & \underline{87.13} & 69.86$_{\pm0.48}$ \\
        ~ & ADFCNN & 72.94 & 53.80 & 91.09 & 55.45 & 50.17 & 56.44 & 57.76 & 85.48 & 85.81& 67.66$_{\pm1.04}$ \\
        ~ & Conformer & \underline{82.84} & 56.44 & 95.05 & 61.06 & 60.73 & 62.05 & \underline{69.31} & \underline{94.06} & 81.52 & \underline{73.67}$_{\pm0.45}$ \\
        ~ & FBCNet & 81.19 & 53.14 & 90.43 & 58.09 & 54.79 & 61.06 & 63.04 & 91.42 & 86.14 & 71.03$_{\pm1.31}$ \\
        ~ & EDPNet & 70.63 & 49.51 & 93.40 & 55.78 & 50.83 & 54.13 & 53.14 & 83.83 & 72.94 & 64.91$_{\pm0.86}$ \\
        \midrule
         \multirow{5}{*}{\makecell{EEG Generalized Models\\(30\% Finetuned)}}
         & BIOT & 61.06 & 54.79 & 58.09 & 59.74 & 59.41 & 52.81 & 51.49 & 63.37 & 56.11 & 57.43$_{\pm0.87}$ \\
        ~ & BENDR & 50.17 & 52.48 & 44.22 & 56.77 & 54.79 & 48.85 & 51.49 & 60.40 & 61.39 & 53.39$_{\pm1.76}$ \\
        ~ & LaBraM & 53.80 & 55.12 & 50.50 & 55.48 & 53.14 & 59.08 & 56.44 & 52.48 & 62.38 & 55.38$_{\pm0.72}$ \\
        ~ & CBraMod & 54.46 & 58.42 & 57.10 & 56.44 & 54.46 & 57.10 & 58.09 & 59.08 & 69.64 & 58.31$_{\pm1.01}$ \\
        ~ & EEGPT & 53.01 & 51.62 & 55.32 & 54.17 & 49.31 & 48.38 & 57.18 & 53.70 & 51.16 & 52.65$_{\pm0.81}$ \\

       \midrule
         \multirow{5}{*}{\makecell{EEG Generalized Models\\(80\% Finetuned)}}
         & BIOT & 62.07 & 57.47 & 86.21 & 64.37 & 64.37 & 64.37 & 62.07 & 88.51 & 75.86 & 69.48$_{\pm1.54}$ \\
        ~ & BENDR & 63.22 & 58.62 & 60.92 & 61.94 & 56.32 & 64.37 & 56.32 & 70.12 & 74.71 & 62.95$_{\pm1.25}$ \\
        ~ & LaBraM & 64.37 & 58.62 & 59.77 & 63.22 & 59.77 & 59.77 & 55.17 & 68.97 & 75.86 & 62.84$_{\pm1.66}$ \\
        ~ & CBraMod & 60.92 & \textbf{64.37} & 63.22 & \underline{66.67} & \underline{64.37} & 64.37 & 63.22 & 63.22 & 70.11 & 64.50$_{\pm0.48}$ \\
        ~ & EEGPT & 56.21 & 53.22 & 59.90 & 40.12 & 51.62 & 56.45 & 55.31 & 55.31 & 55.29 & 53.71$_{\pm0.72}$ \\

        \midrule
        \multirow{1}{*}{MI-FM (30\% Finetuned)}
        & MIRepNet (Ours) & \textbf{92.41} & \underline{61.39} & \textbf{96.04} & \textbf{77.89} & \textbf{76.24} & \textbf{72.28} & \textbf{79.87} & 92.08 & \textbf{87.79} & \textbf{81.77}$_{\pm0.27}$\\
        \bottomrule
    \end{tabular}
\end{table*}

\begin{table*}[htpb]     \centering
\fontsize{7.7}{12}\selectfont %
       \caption{Accuracies (\%) in BNCI2015001. The best accuracies are marked in bold, and the second best by an underline.}  \label{tab:15001}
    \begin{tabular}{w{c}{2.6cm}|w{c}{1.7cm}|*{12}{w{c}{0.52cm}}|w{c}{1.0cm}}   \toprule
        Setting & Approach & S0 & S1 & S2 & S3 & S4 & S5 & S6 & S7 & S8 & S9 & S10 & S11 & Avg. \\
        \midrule
         \multirow{9}{*}{\makecell{EEG Specialist Models\\(30\% Trained)}}
         & CSP+LDA & 93.57 & 95.00 & \underline{94.29} & 82.86 & 86.43 & 66.43 & 80.71 & 62.14 & 52.86 & 59.29 & 82.14 & 50.71 & 75.54 \\
        ~ & ShallowConv & 94.52 & 90.71 & 88.10 & 73.10 & 80.71 & 64.76 & 82.62 & \underline{64.52} & 63.33 & 60.95 & \textbf{93.57} & 51.91 & 75.73$_{\pm0.32}$ \\
        ~ & DeepConv & 73.10 & 58.10 & 63.33 & 51.91 & 67.14 & 48.57 & 73.33 & 47.62 & 56.91 & 59.52 & 86.43 & 52.62 & 61.55$_{\pm0.34}$ \\
        ~ & EEGNet & 95.95 & 94.52 & 93.10 & 81.91 & 81.43 & 63.57 & 82.62 & 55.72 & \underline{65.00} & 64.05 & 82.62 & 52.62 & 76.09$_{\pm0.99}$ \\
        ~ & IFNet & 95.95 & \textbf{95.24} & 93.57 & 86.67 & 78.81 & 68.10 & 81.91 & 57.62 & 62.86 & 64.52 & 88.81 & 53.81 & 77.32$_{\pm0.49}$ \\
        ~ & ADFCNN & 95.71 & \textbf{95.24} & 90.48 & 82.62 & 80.48 & 67.14 & 84.05 & 58.10 & 58.33 & 60.48 & 89.29 & 60.24 & 76.85$_{\pm0.84}$ \\
        ~ & Conformer & 95.71 & 94.29 & 88.57 & 81.67 & 85.95 & 68.57 & 84.52 & 53.57 & 63.33 & 68.33 & 89.05 & 55.71 & \underline{77.44}$_{\pm0.64}$ \\
        ~ & FBCNet & 96.19 & 94.05 & 94.05 & 81.67 & 82.38 & 65.24 & 79.76 & 51.14 & 60.71 & 65.48 & 88.10 & 50.24 & 75.81$_{\pm0.17}$ \\
        ~ & EDPNet & 96.19 & \textbf{95.24} & 91.91 & \textbf{89.05} & \underline{88.57} & \underline{70.71} & 77.62 & 59.29 & 62.62 & 56.91 & 75.48 & 50.48 & 76.17$_{\pm0.58}$ \\
        \midrule
         \multirow{5}{*}{\makecell{EEG Generalized Models\\(30\% Finetuned)}}
         & BIOT & \underline{98.57} & 92.38 & 92.38 & 87.14 & 73.57 & 55.48 & 80.72 & 57.14 & 52.14 & 63.57 & 81.43 & 62.86 & 74.78$_{\pm1.41}$ \\
        ~ & BENDR & 59.52 & 51.67 & 55.00 & 52.62 & 54.29 & 55.71 & 57.38 & 54.29 & 55.95 & 55.95 & 51.91 & 53.33 & 54.80$_{\pm0.12}$ \\
        ~ & LaBraM & 68.10 & 57.62 & 55.48 & 59.05 & 55.24 & 52.14 & 54.05 & 51.43 & 58.10 & 50.71 & 51.67 & 57.38 & 55.91$_{\pm0.28}$ \\
        ~ & CBraMod & 80.71 & 72.62 & 72.86 & 65.48 & 56.43 & 56.19 & 57.14 & 52.14 & 54.29 & 56.43 & 58.81 & 56.19 & 61.61$_{\pm0.51}$ \\
        ~ & EEGPT & 72.62 & 69.29 & 57.62 & 58.57 & 58.57 & 50.95 & 57.14 & 54.29 & 52.86 & 58.10 & 59.52 & 52.62 & 58.51$_{\pm1.52}$ \\

        \midrule
         \multirow{5}{*}{\makecell{EEG Generalized Models\\(80\% Finetuned)}}
         & BIOT & \textbf{99.17} & 90 & 94.17 & \underline{87.5} & 76.67 & 60.00 & \underline{85.83} & 61.67 & 55.83 & \underline{71.67} & 70.83 & 60.83 & 76.18$_{\pm1.25}$ \\
        ~ & BENDR & 68.33 & 63.33 & 68.33 & 67.50 & 66.67 & 60.00 & 53.33 & 55.83 & \textbf{72.50} & 70.83 & 54.17 & 53.33 & 62.85$_{\pm1.42}$ \\
        ~ & LaBraM & 83.33 & 62.50 & 61.67 & 71.67 & 58.33 & 64.17 & 64.17 & 62.50 & \underline{65.00} & 64.17 & 54.17 & \underline{63.33} & 64.58$_{\pm0.90}$ \\
        ~ & CBraMod & 94.17 & 77.50 & 84.17 & 68.33 & 70.00 & 60.83 & 66.67 & 60.00 & 59.17 & 62.50 & 60.83 & \textbf{65.00} & 69.10$_{\pm2.44}$ \\
        ~ & EEGPT & 96.67 & 90.00 & 86.67 & 70.83 & 67.50 & 60.83 & 67.50 & 56.00 & 60.83 & 67.50 & 62.50 & 56.67 & 70.29$_{\pm1.44}$ \\

        \midrule
        \multirow{1}{*}{MI-FM (30\% Finetuned)}
        & MIRepNet (Ours) & 97.62 & 95.00 & \textbf{94.76} & 85.48 & \textbf{92.86} & \textbf{71.67} & \textbf{86.43} & \textbf{69.52} & 63.57 & \textbf{76.90} & \underline{91.43} & 54.76 & \textbf{81.67}$_{\pm0.26}$ \\
        \bottomrule
    \end{tabular}
\end{table*}

\begin{table*}[htpb]     \centering
\fontsize{9}{12}\selectfont %
       \caption{Accuracies (\%) in BNCI2014004. The best accuracies are marked in bold, and the second best by an underline.}  \label{tab:14004}
    \begin{tabular}{w{c}{3.2cm}|w{c}{2.1cm}|*{9}{w{c}{0.69cm}}|w{c}{1.3cm}}   \toprule
        Setting & Approach & S0 & S1 & S2 & S3 & S4 & S5 & S6 & S7 & S8 & Avg. \\
        \midrule
         \multirow{9}{*}{\makecell{EEG Specialist Models\\(30\% Trained)}}
         & CSP+LDA & 82.14 & 50.40 & 53.87 & 97.32 & 82.74 & 79.76 & 63.69 & 85.71 & 82.44 & 75.34 \\
        ~ & ShallowConv & \underline{83.04} & 59.52 & 51.49 & 98.51 & \underline{90.48} & 71.73 & \underline{78.87} & 84.52 & 85.71 & 78.21$_{\pm0.19}$ \\
        ~ & DeepConv & 77.08 & 53.57 & 52.68 & 98.51 & 87.20 & 67.26 & 74.70 & 84.23 & 88.10 & 75.93$_{\pm1.00}$ \\
        ~ & EEGNet & 74.70 & 55.95 & 52.68 & 97.92 & 83.04 & 68.75 & 66.97 & 84.52 & 78.27 & 73.64$_{\pm0.21}$ \\
        ~ & IFNet & 81.55 & 51.98 & 51.19 & 97.62 & 88.10 & 79.76 & 76.49 & 86.31 & \underline{89.58} & 78.06$_{\pm0.84}$ \\
        ~ & ADFCNN & 75.89 & 57.54 & 55.06 & 98.21 & 89.88 & 71.43 & 75.89 & 85.42 & 86.61 & 77.33$_{\pm0.31}$ \\
        ~ & Conformer & 81.25 & 57.14 & 51.49 & 98.21 & \textbf{92.56} & \underline{82.74} & 78.27 & 83.33 & 86.01 & \underline{79.00}$_{\pm0.68}$ \\
        ~ & FBCNet & 79.46 & 50.00 & 51.49 & 98.51 & 87.20 & 81.55 & 73.21 & 86.01 & 84.82 & 76.92$_{\pm0.13}$ \\
        ~ & EDPNet & 68.45 & 55.16 & 53.57 & 98.21 & 85.71 & 81.25 & 75.30 & 83.04 & 82.74 & 75.94$_{\pm1.22}$ \\
        \midrule
         \multirow{5}{*}{\makecell{EEG Generalized Models\\(30\% Finetuned)}}
         & BIOT & 77.68 & 53.17 & 55.95 & 92.26 & 86.01 & 59.52 & 73.21 & 82.44 & 79.46 & 73.30$_{\pm0.56}$ \\
        ~ & BENDR & 59.82 & 54.37 & 58.33 & 64.88 & 56.85 & 53.27 & 57.74 & 55.66 & 55.06 & 57.33$_{\pm0.46}$ \\
        ~ & LaBraM & 57.14 & 60.71 & 57.14 & 57.44 & 61.31 & 57.14 & 55.36 & 53.87 & 53.27 & 57.04$_{\pm0.67}$ \\
        ~ & CBraMod & 65.18 & 53.97 & 55.36 & 97.32 & 58.04 & 61.90 & 56.55 & 87.80 & 65.77 & 66.88$_{\pm4.46}$ \\
        ~ & EEGPT & 58.63 & 57.54 & 55.95 & 56.55 & 50.30 & 59.52 & 57.44 & 57.44 & 51.79 & 56.13$_{\pm0.15}$ \\

       \midrule
         \multirow{5}{*}{\makecell{EEG Generalized Models\\(80\% Finetuned)}}
         & BIOT & 82.29 & 59.72 & 60.42 & 89.58 & 86.46 & 58.33 & 77.08 & \underline{88.54} & 85.42 & 76.43$_{\pm0.14}$ \\
        ~ & BENDR & 60.42 & 55.21 & 53.13 & 73.96 & 56.25 & 65.63 & 61.46 & 64.58 & 62.50 & 61.46$_{\pm1.09}$ \\
        ~ & LaBraM & 59.38 & 59.72 & 61.46 & 59.38 & 62.50 & 60.42 & 66.67 & 61.46 & 67.71 & 62.08$_{\pm1.20}$ \\
        ~ & CBraMod & 73.96 & 61.11 & 60.42 & \textbf{100.0} & 66.67 & 77.08 & 64.58 & \textbf{90.63} & 86.46 & 75.66$_{\pm0.58}$ \\
        ~ & EEGPT & 60.76 & \textbf{65.28} & \underline{61.46} & 64.58 & 63.54 & 60.42 & 69.79 & 56.25 & 59.38 & 62.38$_{\pm2.75}$ \\
        \midrule
        \multirow{1}{*}{MI-FM (30\% Finetuned)}
        & MIRepNet (Ours) & \textbf{85.71} & \underline{61.51} & \textbf{61.61} & \underline{98.81} & 88.99 & \textbf{85.42} & \textbf{79.17} & 88.39 & \textbf{91.67} & \textbf{82.36}$_{\pm0.10}$ \\
        \bottomrule
    \end{tabular}
\end{table*}

\begin{table*}[htpb]     \centering
\fontsize{9}{12}\selectfont %
       \caption{Accuracies (\%) in AlexMI. The best accuracies are marked in bold, and the second best by an underline.}  \label{tab:AlexMI}
    \begin{tabular}{w{c}{3.1cm}|w{c}{2.1cm}|*{8}{w{c}{0.86cm}}|w{c}{1.3cm}}   \toprule
        Setting & Approach & S0 & S1 & S2 & S3 & S4 & S5 & S6 & S7 & Avg. \\
        \midrule
         \multirow{9}{*}{\makecell{EEG Specialist Models\\(30\% Trained)}}
         & CSP+LDA & 42.86 & 60.71 & 60.71 & 64.29 & 67.86 & \textbf{60.71} & 92.86 & 53.57 & 62.95 \\
        ~ & ShallowConv & 38.10 & 46.43 & 60.71 & 55.95 & 50.00 & 51.19 & 84.52 & 61.91 & 56.10$_{\pm0.42}$ \\
        ~ & DeepConv & 53.57 & 50.00 & 50.00 & 61.91 & 52.38 & 39.29 & 45.24 & 57.14 & 51.19$_{\pm0.92}$ \\
        ~ & EEGNet & 41.67 & 53.57 & 53.57 & 67.86 & 46.43 & 42.86 & 69.05 & 48.81 & 52.98$_{\pm5.67}$ \\
        ~ & IFNet & 39.29 & 50.00 & 51.19 & 53.57 & 48.81 & 47.62 & 48.81 & 71.43 & 51.34$_{\pm1.31}$ \\
        ~ & ADFCNN & 47.62 & 44.05 & 61.91 & 55.95 & 53.57 & 53.57 & 86.91 & 71.43 & 59.38$_{\pm3.24}$ \\
        ~ & Conformer & 41.67 & \textbf{66.67} & \underline{72.62} & 59.52 & 63.10 & 46.43 & 85.71 & \underline{72.62} & 63.54$_{\pm2.74}$ \\
        ~ & FBCNet & 40.48 & \underline{64.29} & 70.24 & 66.67 & 57.14 & 40.48 & 61.91 & 67.86 & 58.63$_{\pm1.28}$ \\
        ~ & EDPNet & 39.29 & 61.91 & 60.71 & 60.71 & 46.43 & 42.86 & 51.19 & 42.86 & 50.74$_{\pm1.52}$ \\
        \midrule
         \multirow{5}{*}{\makecell{EEG Generalized Models\\(30\% Finetuned)}}
         & BIOT & 55.95 & 57.14 & 52.38 & 61.91 & 57.14 & 52.38 & 77.38 & 64.29 & 59.82$_{\pm1.09}$ \\
        ~ & BENDR & \textbf{57.14} & 53.57 & 55.95 & 53.57 & 55.95 & 57.14 & 59.52 & 53.57 & 55.80$_{\pm1.67}$ \\
        ~ & LaBraM & 54.76 & 50.00 & 59.52 & 57.14 & 52.38 & 54.76 & 57.14 & 60.71 & 55.80$_{\pm1.09}$ \\
        ~ & CBraMod & 55.95 & 60.71 & 57.14 & 65.48 & 58.33 & \underline{58.33} & 63.10 & \underline{72.62} & 61.46$_{\pm1.05}$ \\
        ~ & EEGPT & 51.19 & 52.38 & 47.62 & 58.33 & 55.95 & \underline{58.33} & 52.38 & 50.00 & 53.27$_{\pm1.28}$ \\

        \midrule
         \multirow{5}{*}{\makecell{EEG Generalized Models\\(80\% Finetuned)}}
         & BIOT & 54.17 & 58.33 & 70.83 & \textbf{70.83} & \underline{70.83} & 50.00 & \underline{95.83} & 62.50 & \underline{66.67}$_{\pm3.21}$ \\
        ~ & BENDR & 54.17 & 58.33 & 62.50 & 66.67 & \underline{70.83} & 50.00 & 79.17 & 70.83 & 64.06$_{\pm1.28}$ \\
        ~ & LaBraM & 54.17 & 58.33 & 66.67 & 58.33 & 54.17 & 54.17 & 54.17 & 62.50 & 57.81$_{\pm2.55}$ \\
        ~ & CBraMod & 54.17 & 59.17 & 58.33 & \textbf{70.83} & 61.67 & \underline{58.33} & 66.67 & 70.83 & 62.50$_{\pm1.17}$ \\
        ~ & EEGPT & 54.17 & 58.33 & 54.17 & 62.50 & 58.33 & 54.17 & 58.33 & 62.50 & 57.81$_{\pm3.38}$ \\
        \midrule
        \multirow{1}{*}{MI-FM (30\% Finetuned)}
        & MIRepNet (Ours) & \underline{55.95} & 59.52 & \textbf{73.81} & \underline{70.24} & \textbf{75.00} & 52.38 & \textbf{96.43} & \textbf{75.00} & \textbf{69.79}$_{\pm0.42}$ \\
        \bottomrule
    \end{tabular}
\end{table*}

\begin{table*}[htpb]     \centering
\fontsize{9}{12}\selectfont %
       \caption{Accuracies (\%) in BNCI2014001-4. The best accuracies are marked in bold, and the second best by an underline.}  \label{tab:14001_4}
    \begin{tabular}{w{c}{3.2cm}|w{c}{2.1cm}|*{9}{w{c}{0.7cm}}|w{c}{1.4cm}}   \toprule
        Setting & Approach & S0 & S1 & S2 & S3 & S4 & S5 & S6 & S7 & S8 & Avg. \\
        \midrule
         \multirow{9}{*}{\makecell{EEG Specialist Models\\(30\% Trained)}}
         & CSP+LDA & \textbf{72.28} & \underline{55.94} & \underline{77.72} & 40.10 & 35.15 & 44.06 & \underline{78.22} & 76.24 & 64.36 & \underline{60.45} \\
        ~ & ShallowConv & 68.65 & 51.82 & 67.33 & 39.60 & 32.67 & 39.77 & 56.93 & 65.68 & \underline{66.67} & 54.35$_{\pm0.41}$ \\
        ~ & DeepConv & 31.52 & 27.23 & 43.89 & 28.38 & 24.59 & 27.39 & 23.74 & 26.57 & 47.03 & 31.37$_{\pm0.10}$ \\
        ~ & EEGNet & 59.41 & 35.81 & 71.45 & 32.01 & 33.33 & 32.84 & 58.09 & 52.31 & 62.38 & 48.63$_{\pm0.51}$ \\
        ~ & IFNet & 70.63 & \textbf{56.27} & 71.78 & 38.61 & 34.98 & 37.95 & 68.48 & 69.97 & \textbf{70.96} & 57.74$_{\pm0.88}$ \\
        ~ & ADFCNN & 64.36 & 50.17 & 67.00 & 39.11 & 34.16 & 32.51 & 63.70 & 59.24 & 63.37 & 52.62$_{\pm0.45}$ \\
        ~ & Conformer & 69.80 & 48.68 & 72.77 & \underline{44.39} & 37.46 & \textbf{45.88} & 71.78 & \underline{79.04} & 61.22 & 59.00$_{\pm0.69}$ \\
        ~ & FBCNet & \textbf{72.28} & 54.95 & 77.39 & 41.09 & 34.32 & 39.27 & 70.46 & 72.61 & 64.52 & 58.54$_{\pm0.59}$ \\
        ~ & EDPNet & 64.03 & 37.79 & 69.14 & 36.47 & 30.53 & 32.84 & 48.68 & 67.66 & 57.92 & 49.45$_{\pm0.80}$ \\
        \midrule
         \multirow{5}{*}{\makecell{EEG Generalized Models\\(30\% Finetuned)}}
         & BIOT & 58.09 & 45.55 & 39.44 & 31.02 & 30.69 & 28.55 & 46.04 & 49.51 & 36.96 & 40.65$_{\pm0.48}$ \\
        ~ & BENDR & 27.72 & 27.56 & 25.74 & 31.68 & 28.71 & 27.72 & 25.58 & 32.18 & 35.81 & 29.19$_{\pm1.04}$ \\
        ~ & LaBraM & 34.49 & 31.35 & 33.04 & 31.02 & 28.71 & 26.90 & 30.03 & 31.35 & 39.11 & 31.78$_{\pm0.41}$ \\
        ~ & CBraMod & 36.63 & 30.69 & 41.09 & 31.85 & 32.34 & 30.53 & 33.50 & 41.09 & 45.38 & 35.90$_{\pm0.38}$ \\
        ~ & EEGPT & 36.14 & 28.38 & 26.05 & 28.71 & 33.17 & 27.39 & 30.03 & 29.54 & 33.99 & 30.38$_{\pm0.78}$ \\

       \midrule
         \multirow{5}{*}{\makecell{EEG Generalized Models\\(80\% Finetuned)}}
         & BIOT & 70.69 & 54.02 & 69.54 & 37.93 & \underline{44.25} & 38.51 & 71.26 & 75.29 & 50.00 & 56.83$_{\pm1.01}$ \\
        ~ & BENDR & 38.51 & 35.06 & 31.61 & 35.06 & 37.36 & 37.36 & 34.48 & 39.08 & 43.10 & 36.85$_{\pm0.94}$ \\
        ~ & LaBraM & 51.72 & 41.38 & 37.36 & 32.18 & 37.93 & 35.63 & 31.03 & 36.78 & 47.13 & 39.02$_{\pm1.41}$ \\
        ~ & CBraMod & 47.13 & 33.33 & 46.55 & 39.66 & 35.63 & 39.66 & 44.83 & 47.13 & 54.02 & 43.10$_{\pm0.63}$ \\
        ~ & EEGPT & 54.02 & 35.06 & 36.78 & 39.08 & 33.33 & 29.31 & 35.63 & 39.08 & 37.93 & 37.80$_{\pm1.35}$ \\    

        \midrule
        \multirow{1}{*}{MI-FM (30\% Finetuned)}
        & MIRepNet (Ours) & \underline{71.78} & 55.12 & \textbf{83.00} & \textbf{52.97} & \textbf{49.34} & \underline{45.21} & \textbf{78.55} & \textbf{81.02} & 60.23 & \textbf{64.14}$_{\pm0.31}$ \\
        \bottomrule
    \end{tabular}
\end{table*}

\begin{table*}[t]     \centering
\fontsize{9}{12}\selectfont %
       \caption{Accuracies (\%) in five downstream datasets. The best accuracies are marked in bold, and the second best by an underline.}  \label{tab:all_exp}
    \begin{tabular}{w{c}{3.3cm}|w{c}{2.0cm}|*{6}{w{c}{1.9cm}}}   \toprule
        Setting & Approach & BNCI2014001 & BNCI2015001 & BNCI2014004 & AlexMI & BNCI2014001-4 \\
        \midrule
         \multirow{9}{*}{\makecell{EEG Specialist Models\\(30\% Trained)}}
         & CSP+LDA & 72.06 & 75.54 & 75.34 & 62.95 & \underline{60.45} \\
        ~ & ShallowConv & 68.94$_{\pm1.15}$ & 75.73$_{\pm0.32}$ & 78.21$_{\pm0.19}$ & 56.10$_{\pm0.42}$ & 54.35$_{\pm0.41}$ \\
        ~ & DeepConv & 60.07$_{\pm1.65}$ & 61.55$_{\pm0.34}$ & 75.93$_{\pm1.00}$ & 51.19$_{\pm0.92}$ & 31.37$_{\pm0.10}$ \\
        ~ & EEGNet & 62.38$_{\pm2.19}$ & 76.09$_{\pm0.99}$ & 73.64$_{\pm0.21}$ & 52.98$_{\pm5.67}$ & 48.63$_{\pm0.51}$ \\
        ~ & IFNet & 69.86$_{\pm0.48}$ & 77.32$_{\pm0.49}$ & 78.06$_{\pm0.84}$ & 51.34$_{\pm1.31}$ & 57.74$_{\pm0.88}$ \\
        ~ & ADFCNN & 67.66$_{\pm1.04}$ & 76.85$_{\pm0.84}$ & 77.33$_{\pm0.31}$ & 59.38$_{\pm3.24}$ & 52.62$_{\pm0.45}$ \\
        ~ & Conformer & \underline{73.67}$_{\pm0.45}$ & \underline{77.44}$_{\pm0.64}$ & \underline{79.00}$_{\pm0.68}$ & 63.54$_{\pm2.74}$ & 59.00$_{\pm0.69}$ \\
        ~ & FBCNet & 71.03$_{\pm1.31}$ & 75.81$_{\pm0.17}$ & 76.92$_{\pm0.13}$ & 58.63$_{\pm1.28}$ & 58.54$_{\pm0.59}$ \\
        ~ & EDPNet & 64.91$_{\pm0.86}$ & 76.17$_{\pm0.58}$ & 75.94$_{\pm1.22}$ & 50.74$_{\pm1.52}$ & 49.45$_{\pm0.80}$ \\

       \midrule
         \multirow{5}{*}{\makecell{EEG Generalized Models\\(80\% Finetuned)}}
         & BIOT & 69.48$_{\pm1.54}$ & 76.18$_{\pm1.25}$ & 76.43$_{\pm0.14}$ & \underline{66.67}$_{\pm3.21}$ & 56.83$_{\pm1.01}$ \\
        ~ & BENDR & 62.95$_{\pm1.25}$ & 62.85$_{\pm1.42}$ & 61.46$_{\pm1.09}$ & 64.06$_{\pm1.28}$ & 36.85$_{\pm0.94}$ \\
        ~ & LaBraM & 62.84$_{\pm1.66}$ & 64.58$_{\pm0.90}$ & 62.08$_{\pm1.20}$ & 57.81$_{\pm2.55}$ & 39.02$_{\pm1.41}$ \\
        ~ & CBraMod  & 64.50$_{\pm0.48}$ & 69.10$_{\pm2.44}$ & 75.66$_{\pm0.58}$ & 62.50$_{\pm1.17}$ & 43.10$_{\pm0.63}$ \\
        ~ & EEGPT & 53.71$_{\pm0.72}$ & 70.29$_{\pm1.44}$ & 62.38$_{\pm2.75}$ & 57.81$_{\pm3.38}$ & 37.80$_{\pm1.35}$ \\
        \midrule
        \multirow{1}{*}{MI-FM (30\% Finetuned)}
        & MIRepNet (Ours) & \textbf{81.77}$_{\pm0.27}$ & \textbf{81.67}$_{\pm0.26}$ & \textbf{82.36}$_{\pm0.10}$ & \textbf{69.79}$_{\pm0.42}$ & \textbf{64.14}$_{\pm0.31}$ \\
        \bottomrule
    \end{tabular}
    \vspace{-0em}
\end{table*}

\section{Experiments}

\subsection{Datasets}

We conducted experiments utilizing various MI datasets from MOABB \cite{Jayaram2018}, summarized in Table~\ref{tab:datasets}. Seven publicly available EEG datasets, \textit{BNCI2014002} \cite{steyrl2016random}, \textit{PhysionetMI} \cite{goldberger2000physiobank}, \textit{Dreyer2023} \cite{dreyer2023large}, \textit{Weibo2014} \cite{yi2014evaluation}, \textit{Zhou2016} \cite{zhou2016fully}, \textit{Lee2019} \cite{lee2019eeg}, and \textit{Cho2017} \cite{cho2017eeg}, were used for pretraining the MIRepNet foundation model. We further evaluated the downstream performance on five independent datasets (47 subjects): \textit{BNCI2014001} \cite{tangermann2012review}, \textit{BNCI2015001} \cite{faller2012autocalibration}, \textit{BNCI2014004} \cite{leeb2007brain}, \textit{AlexMI} \cite{alexandre2006commande}, and \textit{BNCI2014001-4} \cite{tangermann2012review} covering diverse electrode configurations, sampling frequencies, and MI tasks.

\subsection{Experiment Settings}

All EEG signals were uniformly resampled to 250 Hz during preprocessing. For MIRepNet pretraining, the masking ratio $\alpha$ was set to 50\%. The token embedding dimension $D$ was set to 256, and the Transformer encoder consisted of 6 layers with a dropout rate of 0.5. We trained MIRepNet for 100 epochs using the Adam optimizer with a learning rate of $1\times10^{-3}$ and a batch size of 64. During downstream adaptation, models were finetuned on a limited subset of trials (30\% trials in each subject's session, or 12–86 trials) and generally converged within $\sim$10 epochs. All reported results were averaged across three runs with different random seeds.

\subsection{Main Results}

We compared MIRepNet with nine specialist algorithms and five existing EEG generalist models. Specialist models include CSP+LDA \cite{blankertz2007optimizing}, ShallowConv \cite{schirrmeister2017deep}, DeepConv \cite{schirrmeister2017deep}, EEGNet \cite{lawhern2018eegnet}, IFNet \cite{wang2023ifnet}, ADFCNN \cite{tao2023adfcnn}, Conformer \cite{song2022eeg}, FBCNet\cite{mane2021fbcnet} and EDPNet \cite{han2024edpnet}. EEG generalized models include BIOT \cite{yang2023biot}, BENDR \cite{kostas2021bendr}, LaBraM \cite{jiang2024large}, CBraMod \cite{wang2024cbramod} and EEGPT \cite{wang2025eegpt}.

Tables~\ref{tab:14001}-\ref{tab:all_exp} present the main results. Our proposed MIRepNet consistently achieved the highest average performance, outperforming both specialist models and generalist models.

Notably, existing EEG generalist models are pretrained by mixing multiple BCI paradigms. Because the underlying neurophysiology differs substantially across paradigms, these algorithms struggle to learn truly useful representations and therefore depend on heavy post-training with large amounts of target data. For example, on BNCI2014001, adapting a generalist model to subject $i$ typically requires combining data from the other eight subjects to retrain/finetune the network. In contrast, our downstream setting is more stringent (motivation is that the pretrained model should rapidly adapt to any subject): MIRepNet was fine-tuned with only 30\% trials of a single session from the target user ($<$ 30 trials per class) and still achieved precise decoding. In the reported results, generalist baselines were fine-tuned with 80\% of the target session.

\subsection{Ablation Studies}

To quantify the contribution of each component, we compared MIRepNet with two of its variants:

\begin{enumerate}
\item w/o Pre-training, trained on 30\% trials of the target session.
\item w/o Self-supervised, and the pretraining uses only $\mathcal{L}_{\mathrm{cls}}$ (no masked reconstruction).
\end{enumerate}

As shown in Table~\ref{tab:ablation_all}, adding $\mathcal{L}_{\mathrm{rec}}$ or the pretraining process always improved the performance.

\begin{table*}[htpb]     \centering
\vspace{-1em}
\fontsize{7.7}{12}\selectfont %
       \caption{Accuracies (\%) of ablation studies on five downstream tasks. The best accuracies of each dataset are marked in bold.}  \label{tab:ablation_all}
    \begin{tabular}{w{c}{1.9cm}|w{c}{2.0cm}|*{12}{w{c}{0.54cm}}|w{c}{1.0cm}}   \toprule
        Dataset & Approach & S0 & S1 & S2 & S3 & S4 & S5 & S6 & S7 & S8 & S9 & S10 & S11 & Avg. \\
        \midrule
         \multirow{3}{*}{BNCI2014001}
         & w/o Pre-training & 78.22 & \textbf{62.71} & 89.77 & 58.75 & 75.25 & 61.39 & 67.99 & 91.09 & 80.53 & --- & --- & --- & 73.96$_{\pm1.30}$ \\
        ~ & w/o Self-supervised & 86.47 & 50.17 & 94.72 & 72.28 & 63.37 & 63.70 & 71.95 & 88.78 & 77.89 & --- & --- & --- & 74.37$_{\pm0.39}$ \\
        ~ & MIRepNet (All) & \textbf{92.41} & 61.39 & \textbf{96.04} & \textbf{77.89} & \textbf{76.24} & \textbf{72.28} & \textbf{79.87} & \textbf{92.08} & \textbf{87.79} & --- & --- & --- & \textbf{81.77}$_{\pm0.27}$ \\

        \midrule
         \multirow{3}{*}{BNCI2015001}
         & w/o Pre-training & 97.38 & 91.90 & 93.81 & 84.52 & 85.48 & 69.05 & 85.48 & 57.86 & 58.10 & 69.76 & 88.33 & 55.00 & 78.06$_{\pm0.73}$ \\
        ~ & w/o Self-supervised & \textbf{97.86} & 94.05 & 92.86 & 84.76 & 90.00 & \textbf{74.52} & \textbf{87.62} & 63.57 & \textbf{65.71} & 73.33 & 88.33 & \textbf{55.24} & 80.65$_{\pm0.95}$ \\
        ~ & MIRepNet (All) & 97.62 & \textbf{95.00} & \textbf{94.76} & \textbf{85.48} & \textbf{92.86} & 71.67 & 86.43 & \textbf{69.52} & 63.57 & \textbf{76.90} & \textbf{91.43} & 54.76 & \textbf{81.67}$_{\pm0.26}$ \\

        \midrule
         \multirow{3}{*}{BNCI2014004}
         & w/o Pre-training & 75.30 & 48.02 & 56.55 & 97.92 & 85.42 & 80.06 & 72.32 & 83.93 & 83.04 & --- & --- & --- & 75.84$_{\pm1.14}$ \\
        ~ & w/o Self-supervised & 74.11 & 57.14 & 53.27 & 97.62 & 83.63 & 82.74 & 72.62 & 87.20 & 87.80 & --- & --- & --- & 77.35$_{\pm0.81}$ \\
        ~ & MIRepNet (All) & \textbf{85.71} & \textbf{61.51} & \textbf{61.61} & \textbf{98.81} & \textbf{88.99} & \textbf{85.42} & \textbf{79.17} & \textbf{88.39} & \textbf{91.67} & --- & --- & --- & \textbf{82.36}$_{\pm0.10}$ \\

        \midrule
         \multirow{3}{*}{AlexMI}
         & w/o Pre-training & 47.62 & 59.52 & 70.24 & \textbf{79.76} & 59.52 & \textbf{55.95} & 91.67 & 66.67 & --- & --- & --- & --- & 66.37$_{\pm2.92}$ \\
        ~ & w/o Self-supervised & 52.38 & \textbf{63.10} & 72.62 & 63.10 & 66.67 & 50.00 & 92.86 & \textbf{78.57} & --- & --- & --- & --- & 67.41$_{\pm1.31}$ \\
        ~ & MIRepNet (All) & \textbf{55.95} & 59.52 & \textbf{73.81} & 70.24 & \textbf{75.00} & 52.38 & \textbf{96.43} & 75.00 & --- & --- & --- & --- & \textbf{69.79}$_{\pm0.42}$ \\

        \midrule
         \multirow{3}{*}{BNCI2014001-4}
         & w/o Pre-training & 65.02 & 51.49 & 75.58 & 47.36 & 33.83 & 37.62 & 69.80 & 78.55 & 56.11 & --- & --- & --- & 57.26$_{\pm0.63}$ \\
        ~ & w/o Self-supervised & 69.14 & 47.85 & 82.84 & 46.53 & 36.96 & 43.89 & 77.72 & 74.92 & 56.27 & --- & --- & --- & 59.57$_{\pm0.55}$ \\
        ~ & MIRepNet (All) & \textbf{71.78} & \textbf{55.12} & \textbf{83.00} & \textbf{52.97} & \textbf{49.34} & \textbf{45.21} & \textbf{78.55} & \textbf{81.02} & \textbf{60.23} & --- & --- & --- & \textbf{64.14}$_{\pm0.31}$ \\
        \bottomrule
    \end{tabular}
\end{table*}

\begin{figure*}[t]
\centering
\subfigure[BNCI2014001]{\includegraphics[width=0.2\linewidth]{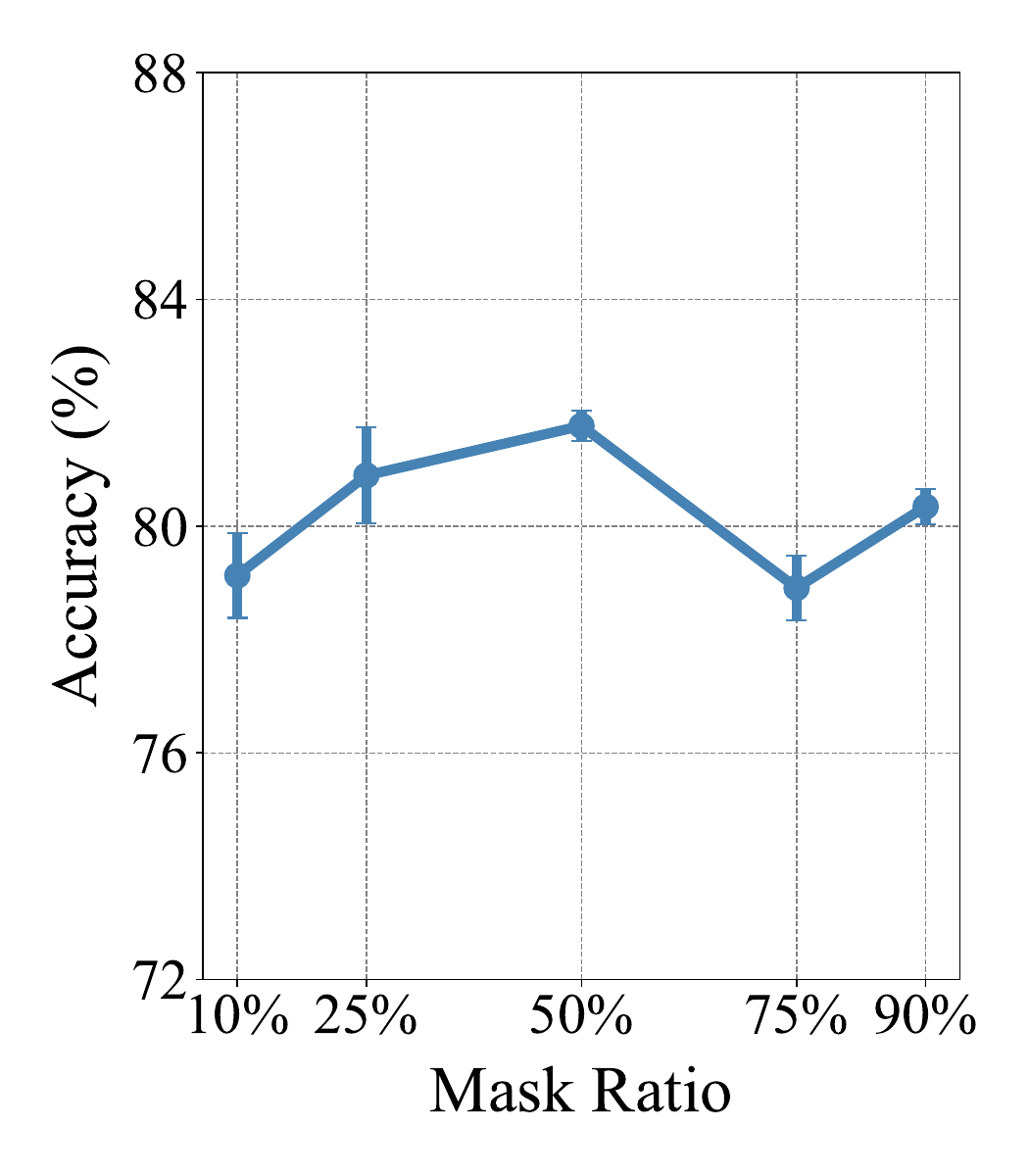}}\hfill
\subfigure[BNCI2015001]{\includegraphics[width=0.2\linewidth]{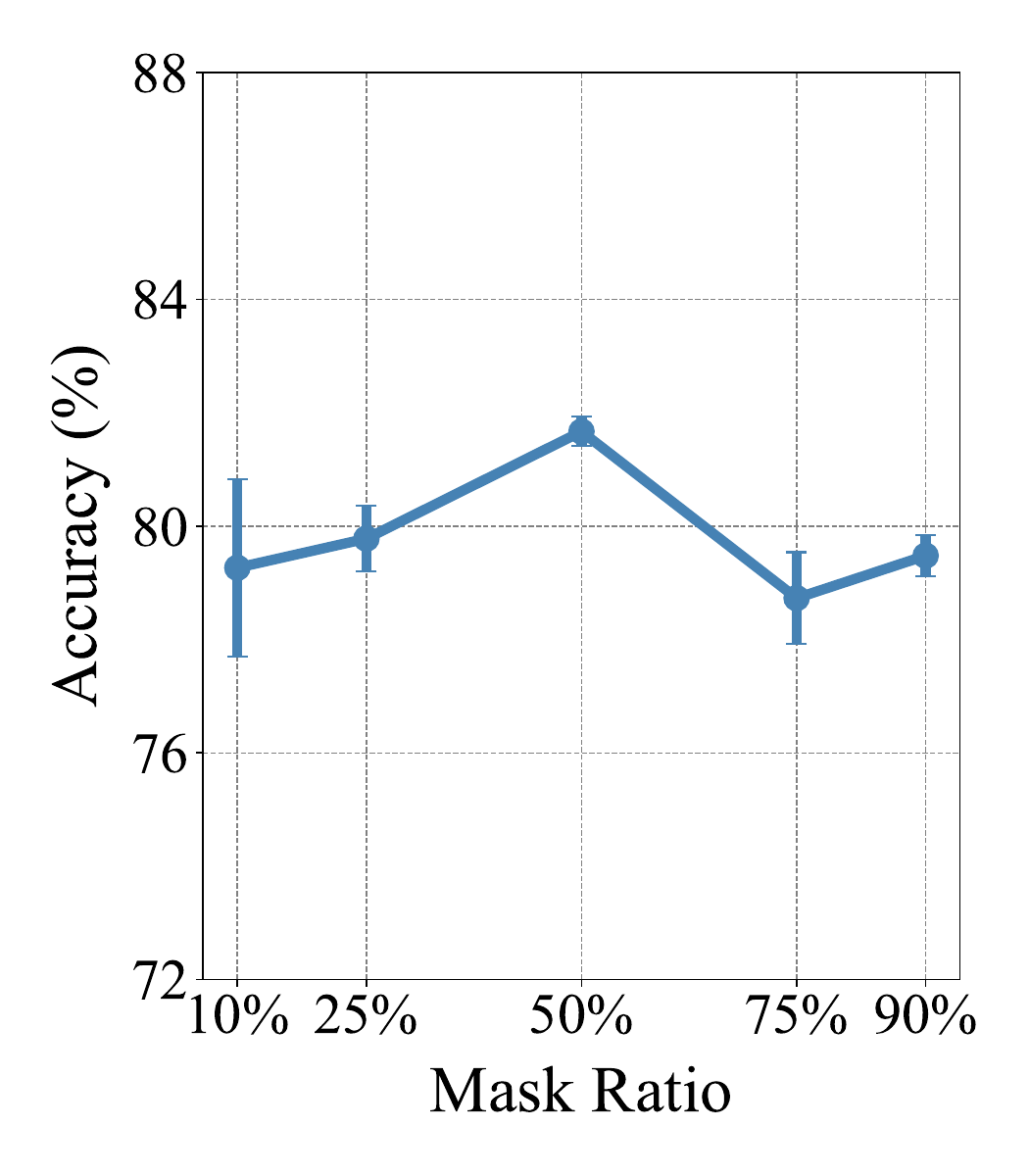}}\hfill
\subfigure[BNCI2014004]{\includegraphics[width=0.2\linewidth]{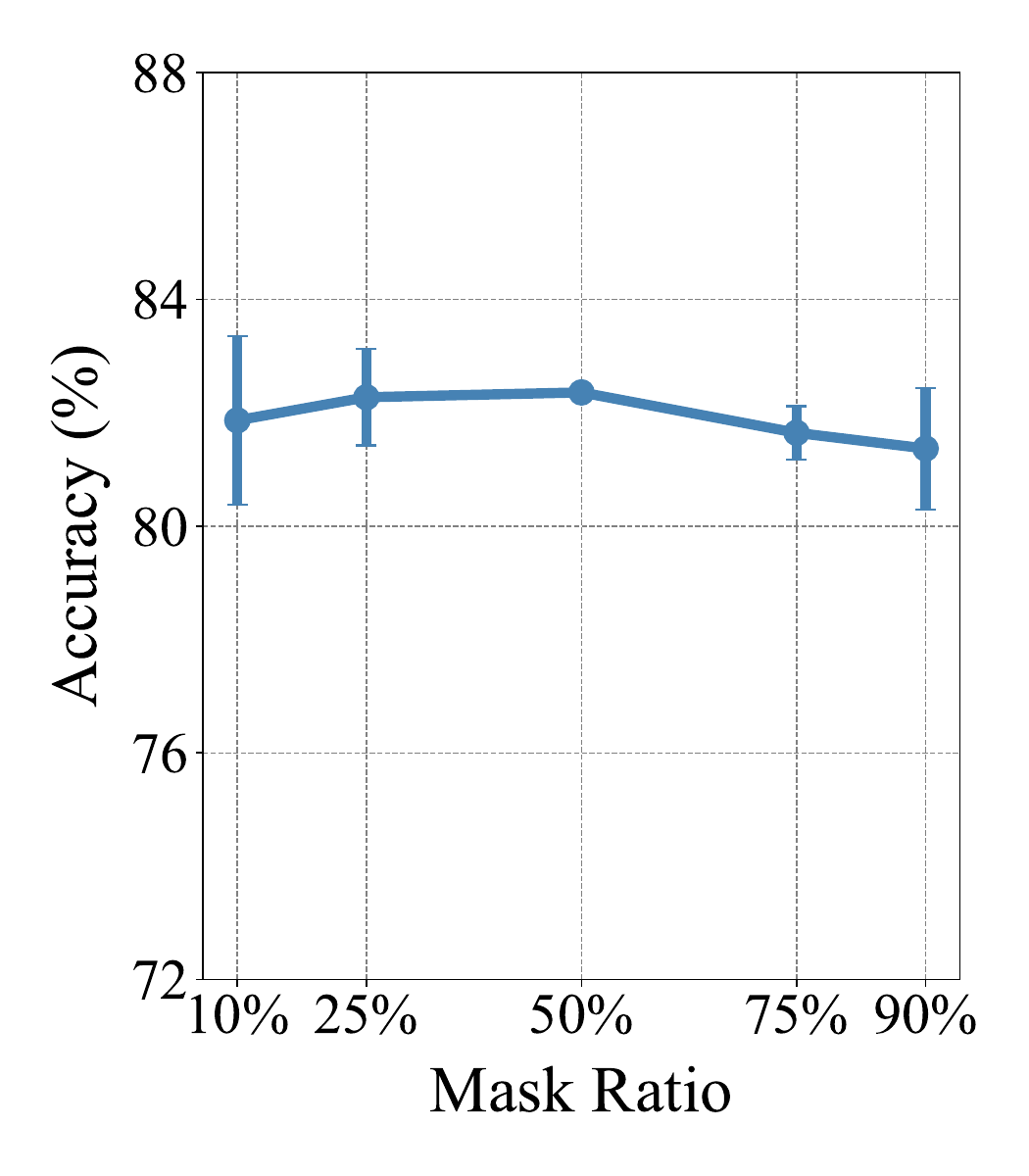}}\hfill
\subfigure[AlexMI]{\includegraphics[width=0.2\linewidth]{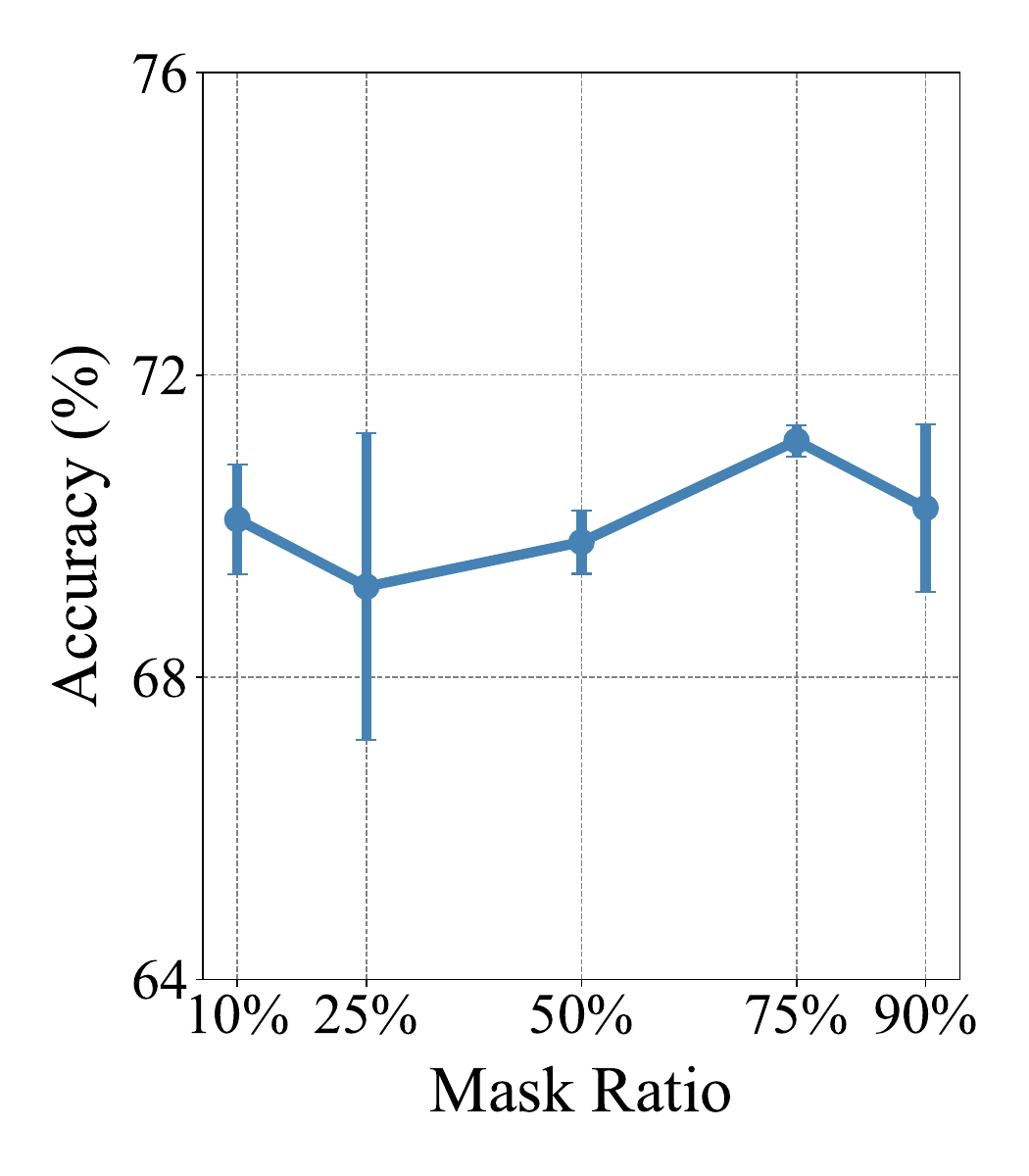}}\hfill
\subfigure[BNCI2014001-4]{\includegraphics[width=0.2\linewidth]{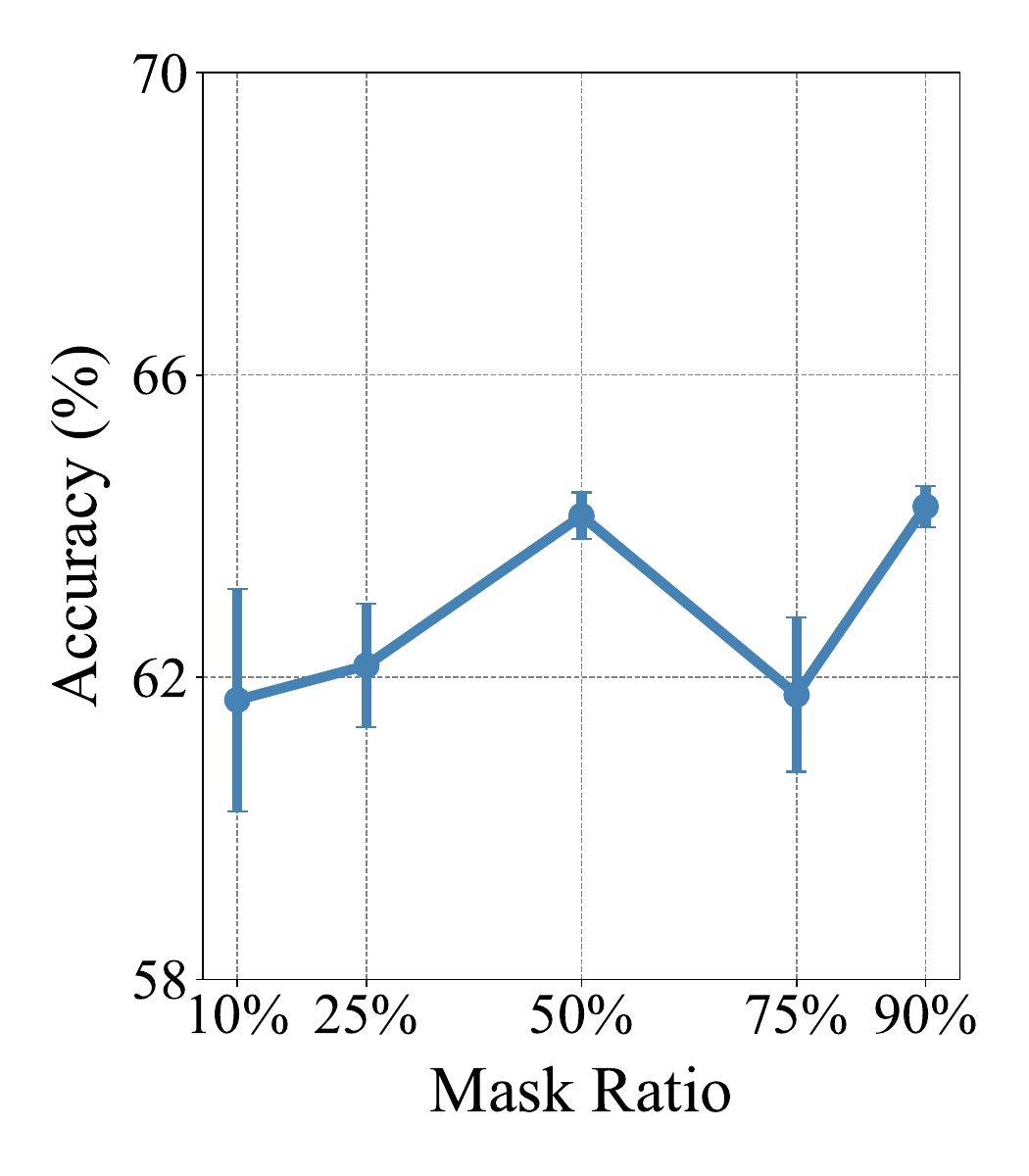}}\hfill
\caption{Sensitivity to mask ratio on three downstream MI datasets.}
\label{fig:mask_sensitivity}
\end{figure*}

\subsection{Analysis of the Mask Ratio}

We varied the masking ratio $\alpha$ on downstream datasets (Fig.~\ref{fig:mask_sensitivity}) and observed only minor fluctuations in accuracy, indicating that MIRepNet is robust to this hyperparameter. The performance peaked around $\alpha=0.5$ and remained competitive for nearby values, confirming that the hybrid objective can effectively exploit both visible and reconstructed tokens. Detailed results are shown in Table~\ref{tab:14004-mask}-~\ref{tab:14001_4-mask}.

\begin{table*}[htpb]     \centering
\fontsize{10}{13}\selectfont %
       \caption{Accuracies (\%) of varying mask ratios in BNCI2014004. The best accuracies are marked in bold.}  \label{tab:14004-mask}
    \begin{tabular}{w{c}{1.8cm}|*{9}{w{c}{1.0cm}}|w{c}{2.0cm}}   \toprule
        Mask Ratio & S0 & S1 & S2 & S3 & S4 & S5 & S6 & S7 & S8 & Avg. \\
        \midrule
         \multirow{5}{*}
        ~ 10\% & 87.80 & 59.13 & 61.31 & \textbf{99.11} & \textbf{89.29} & \textbf{86.90} & 77.38 & 86.61 & 89.29 & 81.87$_{\pm1.49}$ \\
        ~ 25\% & \textbf{88.10} & 58.33 & 62.80 & 98.81 & \textbf{89.29} & 84.52 & \textbf{83.04} & 86.90 & 88.69 & 82.28$_{\pm0.85}$ \\
        ~ 50\% & 85.71 & \textbf{61.51} & 61.61 & 98.81 & 88.99 & 85.42 & 79.17 & \textbf{88.39} & \textbf{91.67} & \textbf{82.36}$_{\pm0.10}$ \\
        ~ 75\% & 85.12 & 59.52 & \textbf{63.69} & 98.51 & 87.50 & \textbf{86.90} & 77.38 & 87.20 & 88.99 & 81.65$_{\pm0.47}$ \\
        ~ 90\% & 85.71 & 59.13 & 61.31 & \textbf{99.11} & 86.01 & 85.12 & 80.95 & 87.20 & 87.80 & 81.37$_{\pm1.07}$ \\
        \bottomrule
    \end{tabular}
\end{table*}

\begin{table*}[htpb]     \centering
\fontsize{10}{13}\selectfont %
       \caption{Accuracies (\%) of varying mask ratios in AlexMI. The best accuracies are marked in bold.}  \label{tab:AlexMI-mask}
    \begin{tabular}{w{c}{1.9cm}|*{8}{w{c}{1.15cm}}|w{c}{2.1cm}}   \toprule
        Mask Ratio & S0 & S1 & S2 & S3 & S4 & S5 & S6 & S7 & Avg. \\
        \midrule
         \multirow{9}{*}
        ~ 10\% & 50.00 & 64.29 & 73.81 & 65.48 & \textbf{83.33} & \textbf{52.38} & 91.67 & \textbf{79.76} & 70.09$_{\pm0.73}$ \\
        ~ 25\% & 57.14 & 63.10 & 73.81 & 70.24 & 71.43 & \textbf{52.38} & 94.05 & 71.43 & 69.20$_{\pm2.03}$ \\
        ~ 50\% & 55.95 & 59.52 & 73.81 & 70.24 & 75.00 & \textbf{52.38} & \textbf{96.43} & 75.00 & 69.79$_{\pm0.42}$ \\
        ~ 75\% & \textbf{59.52} & 61.90 & \textbf{79.76} & 65.48 & 77.38 & 48.81 & \textbf{96.43} & \textbf{79.76} & \textbf{71.13}$_{\pm0.21}$ \\
        ~ 90\% & 47.62 & \textbf{69.05} & 67.86 & \textbf{76.19} & 78.57 & 51.19 & 92.86 & 78.57 & 70.24$_{\pm1.11}$ \\
        \bottomrule
    \end{tabular}
\end{table*}

\begin{table*}[htpb]     \centering
\fontsize{10}{13}\selectfont %
       \caption{Accuracies (\%) of varying mask ratios in BNCI2014001. The best accuracies are marked in bold.}  \label{tab:14001-mask}
    \begin{tabular}{w{c}{1.8cm}|*{9}{w{c}{1.0cm}}|w{c}{2.0cm}}   \toprule
        Mask Ratio & S0 & S1 & S2 & S3 & S4 & S5 & S6 & S7 & S8 & Avg. \\
        \midrule
         \multirow{5}{*}
        ~ 10\% & 90.43 & 58.09 & 95.05 & 73.93 & 75.91 & 68.98 & 80.20 & 91.42 & 78.22 & 79.13$_{\pm0.75}$ \\
        ~ 25\% & 87.46 & \textbf{64.69} & 96.04 & \textbf{79.21} & 73.60 & 69.31 & \textbf{87.46} & 91.42 & 78.88 & 80.89$_{\pm0.85}$ \\
        ~ 50\% & 92.41 & 61.39 & 96.04 & 77.89 & \textbf{76.24} & \textbf{72.28} & 79.87 & 92.08 & \textbf{87.79} & \textbf{81.77}$_{\pm0.27}$ \\
        ~ 75\% & 88.78 & 57.10 & 96.37 & 75.91 & 70.30 & 68.65 & 80.20 & 90.43 & 82.51 & 78.91$_{\pm0.57}$ \\
        ~ 90\% & \textbf{93.07} & 63.70 & \textbf{97.69} & 74.92 & 72.61 & 64.69 & 80.53 & \textbf{92.74} & 83.17 & 80.35$_{\pm0.32}$ \\
        \bottomrule
    \end{tabular}
\end{table*}

\begin{table*}[htpb]     \centering
\fontsize{9}{12}\selectfont %
       \caption{Accuracies (\%) of varying mask ratios in BNCI2015001. The best accuracies are marked in bold.}  \label{tab:15001-mask}
    \begin{tabular}{w{c}{1.6cm}|*{12}{w{c}{0.68cm}}|w{c}{1.8cm}}   \toprule
        Mask Ratio & S0 & S1 & S2 & S3 & S4 & S5 & S6 & S7 & S8 & S9 & S10 & S11 & Avg. \\
        \midrule
         \multirow{5}{*}
         ~ 10\% & 95.71 & 93.81 & 92.14 & 84.52 & 91.43 & 73.57 & 83.33 & 65.24 & \textbf{66.19} & 71.43 & 83.33 & 50.48 & 79.27$_{\pm1.56}$ \\
        ~ 25\% & 97.14 & \textbf{95.24} & \textbf{94.76} & 83.33 & 89.52 & 70.00 & 83.33 & 68.10 & 61.67 & 70.71 & 89.05 & 54.52 & 79.78$_{\pm0.58}$ \\
        ~ 50\% & \textbf{97.62} & 95.00 & \textbf{94.76} & \textbf{85.48} & \textbf{92.86} & 71.67 & \textbf{86.43} & 69.52 & 63.57 & \textbf{76.90} & \textbf{91.43} & \textbf{54.76} & \textbf{81.67}$_{\pm0.26}$ \\
        ~ 75\% & 96.90 & 94.29 & 93.33 & 83.10 & 89.76 & 72.14 & 76.67 & 67.14 & 61.67 & 67.14 & 89.05 & 53.57 & 78.73$_{\pm0.81}$ \\
        ~ 90\% & 96.67 & 92.86 & \textbf{94.76} & 84.29 & 90.95 & \textbf{74.29} & 82.38 & \textbf{70.48} & 63.81 & 66.90 & 87.62 & 48.81 & 79.48$_{\pm0.36}$ \\
        \bottomrule
    \end{tabular}
\end{table*}

\begin{table*}[htpb]     \centering
\fontsize{10}{13}\selectfont %
       \caption{Accuracies (\%) of varying mask ratios in BNCI2014001-4. The best accuracies are marked in bold.}  \label{tab:14001_4-mask}
    \begin{tabular}{w{c}{1.8cm}|*{9}{w{c}{1.0cm}}|w{c}{2.0cm}}   \toprule
        Mask Ratio & S0 & S1 & S2 & S3 & S4 & S5 & S6 & S7 & S8 & Avg. \\
        \midrule
         \multirow{9}{*}
        ~ 10\% & 71.95 & 50.33 & 81.35 & 46.37 & 45.54 & 43.40 & 79.54 & 77.56 & 59.24 & 61.70$_{\pm1.47}$ \\
        ~ 25\% & 68.98 & 49.34 & 81.19 & \textbf{52.97} & \textbf{51.49} & 41.91 & 78.55 & 76.90 & 58.09 & 62.16$_{\pm0.82}$ \\
        ~ 50\% & 71.78 & 55.12 & 83.00 & \textbf{52.97} & 49.34 & \textbf{45.21} & 78.55 & \textbf{81.02} & 60.23 & 64.14$_{\pm0.31}$ \\
        ~ 75\% & 67.66 & 55.28 & 81.85 & 49.17 & 46.86 & 43.23 & 80.20 & 72.28 & 59.41 & 61.77$_{\pm1.02}$ \\
        ~ 90\% & \textbf{74.75} & \textbf{57.59} & 83.50 & 49.01 & 47.03 & 44.55 & \textbf{81.52} & 76.07 & \textbf{64.36} & \textbf{64.26}$_{\pm0.27}$ \\

        \bottomrule
    \end{tabular}
\end{table*}

\section{Discussion}
\subsection{Neurological Principles of MI}

MI signals are associated with the phenomena of event-related desynchronization (ERD) and event-related synchronization (ERS). Specifically, when a subject imagines performing a movement, there is a decrease in the power of specific frequency bands (typically in the alpha and beta bands) in the brain regions associated with the imagined movement (the corresponding frequencies and their effects on behavior are summarized in Table \ref{tab:frequency_bands}). This reduction in power is called event-related desynchronization (ERD) and is typically observed over the sensorimotor cortex, indicating a state of cortical activation. Conversely, if there is no movement imagery, certain brain areas may exhibit an increase in the power of these frequency bands, known as event-related synchronization (ERS). ERD is commonly observed during MI tasks, reflecting the mental preparation or intention to perform a motor action, whereas ERS may be associated with rest or a lack of motor activity \cite{maeder2012pre}.

\begin{table}[htbp]
\centering
\caption{Frequency bands and their characteristics in MI paradigm}
\label{tab:frequency_bands}
\renewcommand{\arraystretch}{1.5} 
\begin{tabular}{c|c|l}
\toprule
\textbf{Band} & \textbf{Range (Hz)} & \textbf{Associated Regions} \\ \hline
$\delta$ & 0.5--4  & Deep sleep, unconscious states \\ \hline
$\theta$ & 4--8    & Relaxation, motor imagery \\ \hline
$\alpha$ & 8--13   & Relaxation, motor imagery \\ \hline
$\beta$  & 13--30  & Motor control \\ \hline
$\gamma$ & 30--45  & Higher cognition \\ 
\bottomrule
\end{tabular}
\end{table}

The influence of the $\alpha$ and $\beta$ frequency bands on MI is most prominent in the sensorimotor cortex. When a person imagines performing a motor task, the corresponding cortical areas in the brain exhibit a decrease in $\alpha$ and $\beta$ band power, a phenomenon associated with motor activity, referred to as ERD. On the other hand, if not performing MI, the brain's $\alpha$ and $\beta$ waves exhibit an increase in power, referred to as ERS. Notably, higher power in the $\beta$ band indicates more pronounced synchronization, and the corresponding task-related brain activity is observed over the contralateral hemisphere. MI tasks involving left-hand and right-hand movements typically show ERD over the C4 and C3 regions, respectively.

\begin{figure}[htbp]         \centering
\includegraphics[width=\linewidth,clip]{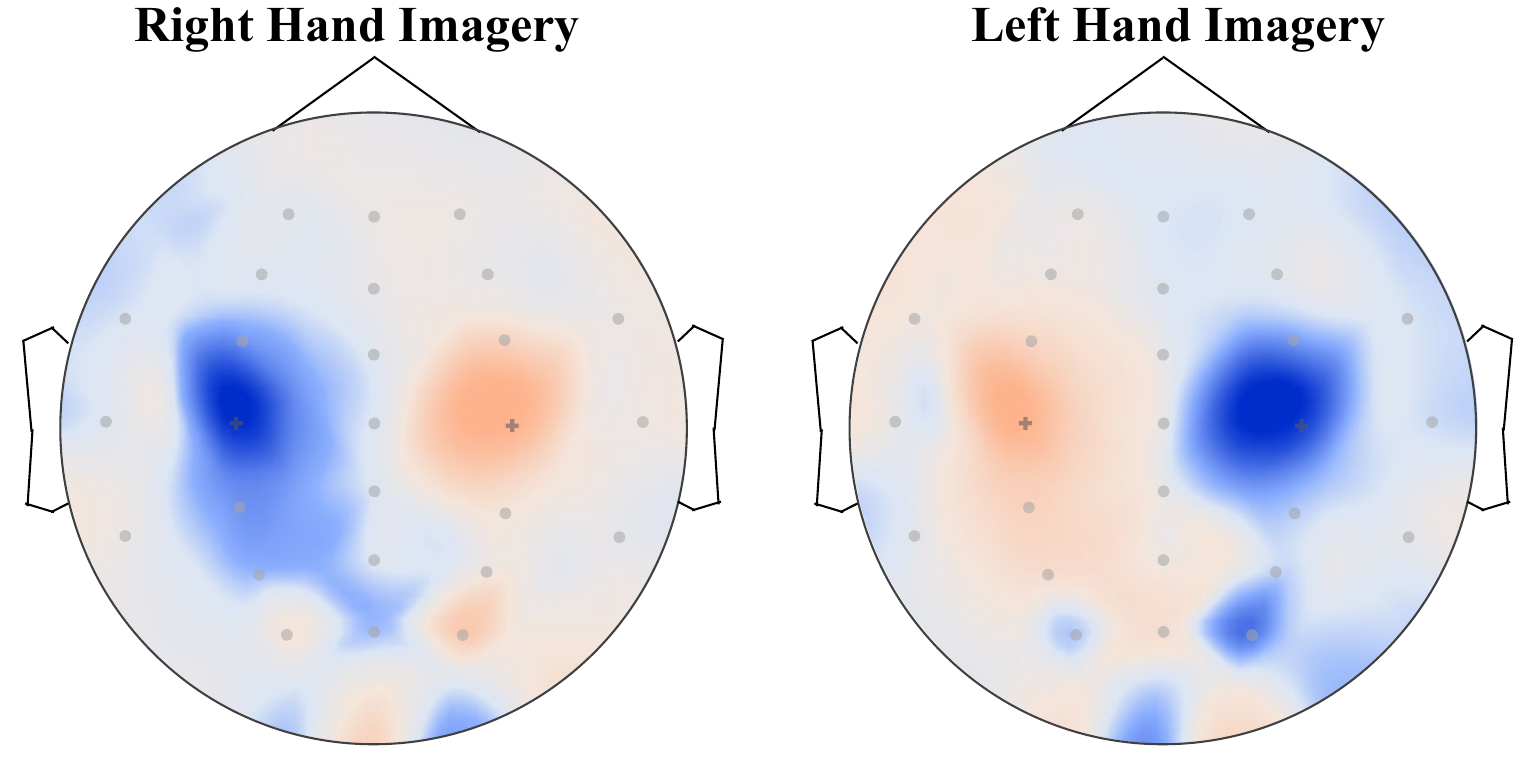}
\caption{Scalp topographies of SMR power changes during motor imagery of the left and right hands. The left panel shows spectral power decreases (blue) predominantly over the left hemisphere during right hand imagery, while the right panel shows power decreases (blue) over the right hemisphere during left hand imagery. The color bar indicates relative amplitude change in the SMR band, with blue denoting power attenuation and red denoting power increase.} \label{fig:SMR}
\end{figure}

Fig.~\ref{fig:SMR} depicts the phenomenon, ERD is observed in the right hemisphere during left-hand imagery and in the left hemisphere during right-hand imagery. These findings are fundamental to BCI systems that decode movement imagery signals from different limbs based on these cortical signatures. Guided by neurophysiological knowledge, we propose a channel template covering frontal‑central (FC), central (C), centro‑parietal (CP), and temporal (T) sites to capture the most informative regions.

\subsection{Datasets Description}

The pretraining and downstream datasets used in this work are summarized below:
\begin{enumerate}
\item BNCI2014002 includes EEG data from 13 participants performing sustained MI of the right hand and feet. The session consists of eight runs, with 50 trials per class for training and 30 trials for validation. EEG was recorded at 512 Hz from 15 electrodes, including C3, Cz, and C4, with a biosignal amplifier and active Ag/AgCl electrodes.
\item PhysionetMI includes over 1500 one- and two-minute EEG recordings from 109 volunteers performing MI tasks. EEG was recorded with 64 channels using the BCI2000 system.
\item Dreyer2023 datatset concatenates Dreyer2023A/B/C and contains 87 subjects. Each recording includes open/closed‑eyes baselines and six MI runs (first two for system acquisition, remaining four for user training), 40 trials per run. Left-/right‑hand MI was recorded with 27 electrodes at 512 Hz.
\item Weibo2014 includes EEG data from 10 subjects recorded with 60 electrodes. It consists of seven mental tasks, including simple and compound limb MI tasks (left hand, right hand, feet, and combinations), and a rest state.
\item Zhou2016 includes EEG data from 4 subjects performing three MI tasks: left hand, right hand, and feet. Each subject participated in three sessions, with each session consisting of two runs of 75 trials (25 trials per class). 
\item Lee2019 includes EEG data recorded from 62 channels at 1,000 Hz using a BrainAmp amplifier, which involved MI tasks for left and right hand grasping, with 100 trials per session. The EEG channels were referenced to the nasion and grounded to AFz.
\item Cho2017 includes EEG data from 52 subjects (19 females, mean age 24.8 ± 3.86 years) performing MI tasks for the left and right hands. EEG was recorded at 512 Hz from 64 channels using the Biosemi ActiveTwo system, with a 10-10 system montage.
\item BNCI2014001 contains EEG data from 9 subjects performing four MI tasks: left hand, right hand, both feet, and tongue. Each subject participated in two sessions, with each session consisting of 6 runs, yielding a total of 288 trials per session.
\item BNCI2015001 contains EEG data from subjects performing sustained MI of the right hand and both feet. The data were recorded at 512 Hz using 15 electrodes, including C3, Cz, and C4, with a bandpass filter between 0.5 and 100 Hz and a notch filter at 50 Hz.
\item BNCI2014004 includes EEG data from 9 right-handed subjects, who performed two MI tasks: left hand and right hand. Each subject participated in five sessions, with the first two for screening without feedback and the last three with feedback. The data was recorded with three bipolar EEG channels (C3, Cz, C4) at 250 Hz and included 120 trials per subject for each MI category.
\item  AlexMI contains EEG recordings from 8 subjects, performing 2 task of motor imagination (right hand, feet or rest). Data have been recorded at 512Hz with 16 wet electrodes.
\end{enumerate}

\subsection{Effectiveness of EA}

\begin{figure}[htbp]
\centering
\begin{minipage}[b]{0.47\linewidth}
  \centering
  \includegraphics[width=\linewidth,clip]{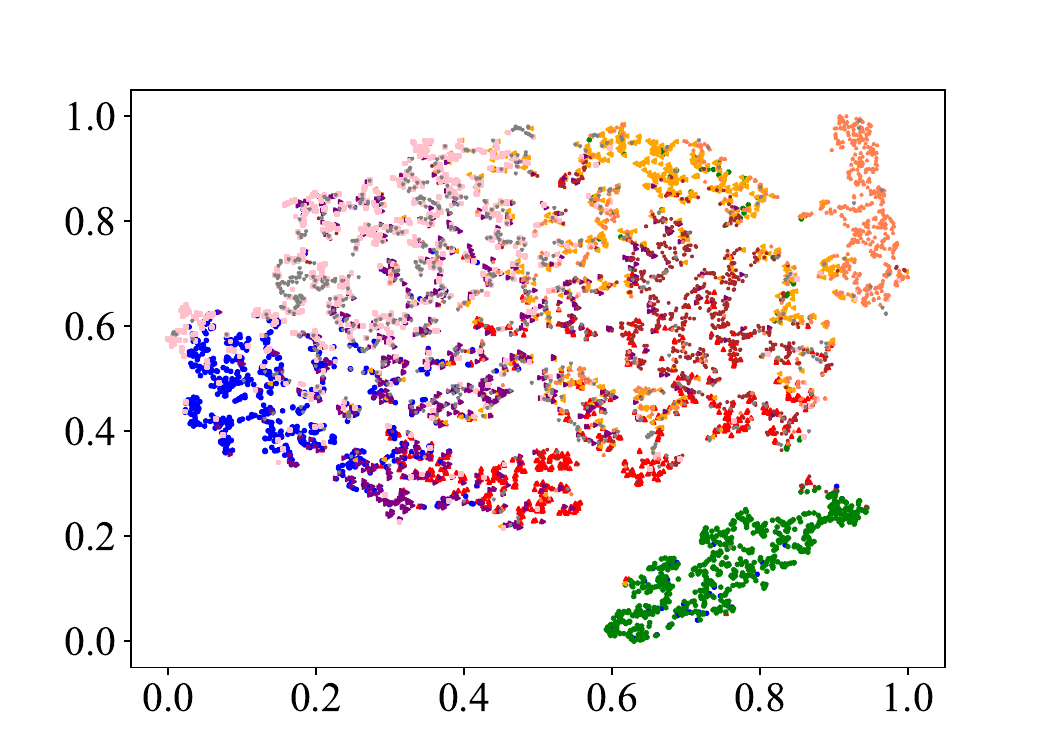}\\[-2pt]
  (a)
\end{minipage}\hfill
\begin{minipage}[b]{0.53\linewidth}
  \centering
  \raisebox{0.4mm}{\includegraphics[width=\linewidth]{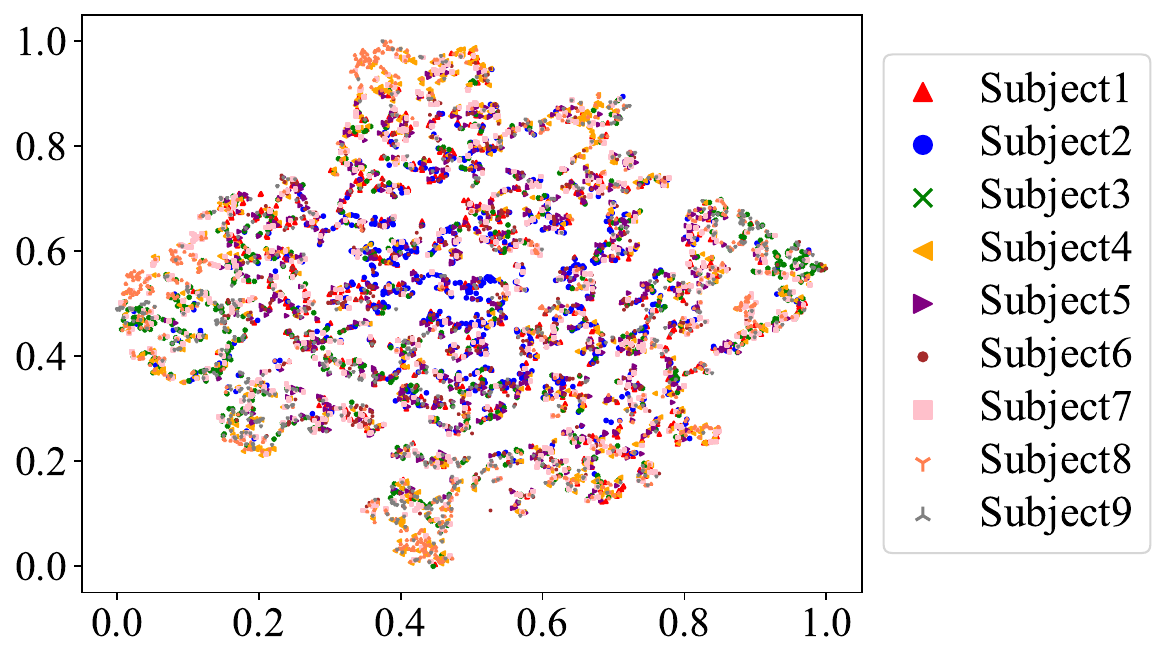}}\\[-2pt] 
  \makebox[\linewidth][l]{\hspace*{0.36\linewidth}(b)}
\end{minipage}
\caption{$t$-SNE visualization of the data in BNCI2014004. (a) Before EA; (b) After EA. Different colors represent trials from different subjects.}
\label{fig:sEA}
\end{figure}

During high‑quality data construction, we apply EA to the channel template-unified EEG so that each subject’s trials share an identity covariance matrix (i.e., second‑order statistics are matched). To visualize its effect, we project the data with $t$‑distributed stochastic neighbor embedding ($t$‑SNE). As shown in Fig.~\ref{fig:sEA}, trials from different subjects are mapped into a common distribution space.

\subsection{Rapid Calibration}  

\begin{figure}[htbp]
\centering
\begin{minipage}[b]{0.5\linewidth}
  \centering
  \raisebox{0.4mm}{\includegraphics[width=\linewidth]{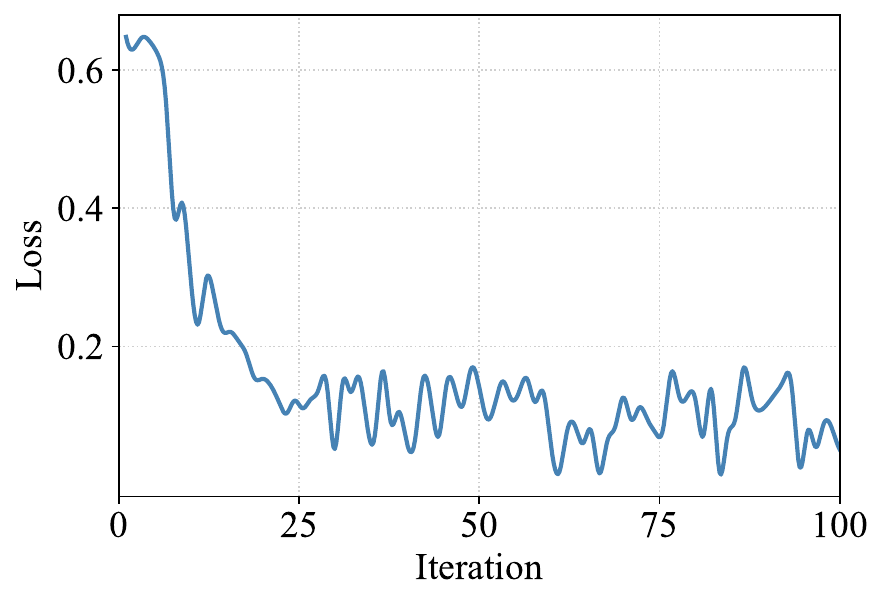}}\\[-2pt] 
  \makebox[\linewidth][l]{\hspace*{0.50\linewidth}(a)}
\end{minipage}\hfill
\begin{minipage}[b]{0.5\linewidth}
  \centering
  \raisebox{0.4mm}{\includegraphics[width=\linewidth]{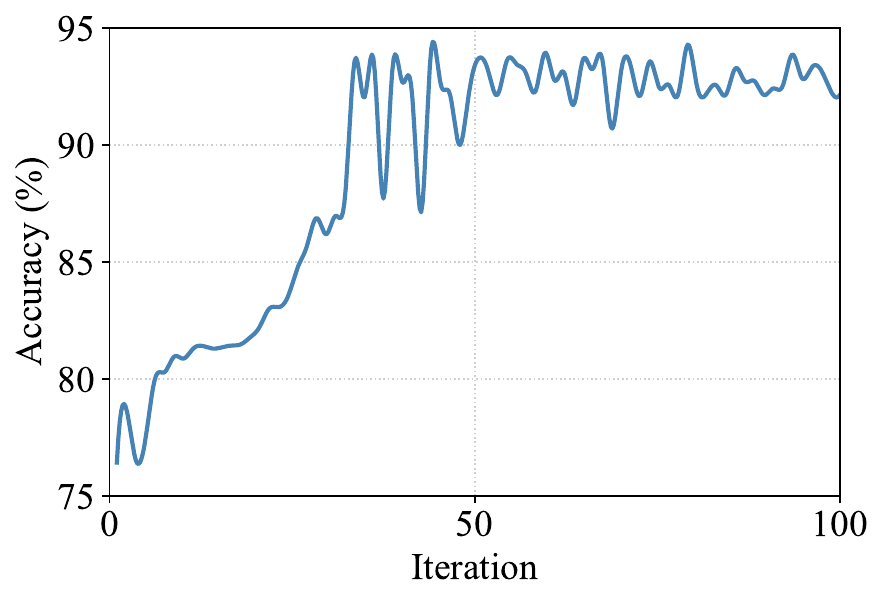}}\\[-2pt] 
  \makebox[\linewidth][l]{\hspace*{0.50\linewidth}(b)}
\end{minipage}
\caption{Test loss and accuracy curves during subject‐specific fine‐tuning of RepMI.}
\label{fig:curve}
\end{figure}

A key benefit of RepMI is its ability to adapt rapidly to downstream subjects and tasks with minimal data.  As shown in Fig.~\ref{fig:curve}, when fine‑tuned on only 30\% of a new subject’s trials, both loss and accuracy converge to near‑peak levels within approximately 10 epochs of training.  This fast convergence demonstrates RepMI’s practicality for low-data, rapid calibration in motor imagery BCI applications.  

\section{Conclusion}

In this paper, we introduce MIRepNet, the first EEG foundation model specifically designed for the MI paradigm. A high-quality EEG data pipeline was developed, featuring a neurophysiologically informed channel template that aligns heterogeneous EEG electrode layouts into a unified spatial framework. Furthermore, an efficient pretraining strategy combining self-supervised masked token reconstruction and supervised MI classification was proposed, enabling rapid adaptation to new subjects and tasks via minimal downstream fine-tuning. Extensive evaluations on five downstream MI tasks encompassing 47 subjects demonstrated the efficacy and robustness of MIRepNet, consistently outperforming state-of-the-art specialist and generalist EEG models. Our results underscore the significant advantage and practical necessity of paradigm-specific EEG foundation models.



\begin{thebibliography}{10}
\providecommand{\url}[1]{#1}
\csname url@samestyle\endcsname
\providecommand{\newblock}{\relax}
\providecommand{\bibinfo}[2]{#2}
\providecommand{\BIBentrySTDinterwordspacing}{\spaceskip=0pt\relax}
\providecommand{\BIBentryALTinterwordstretchfactor}{4}
\providecommand{\BIBentryALTinterwordspacing}{\spaceskip=\fontdimen2\font plus
\BIBentryALTinterwordstretchfactor\fontdimen3\font minus \fontdimen4\font\relax}
\providecommand{\BIBforeignlanguage}[2]{{%
\expandafter\ifx\csname l@#1\endcsname\relax
\typeout{** WARNING: IEEEtran.bst: No hyphenation pattern has been}%
\typeout{** loaded for the language `#1'. Using the pattern for}%
\typeout{** the default language instead.}%
\else
\language=\csname l@#1\endcsname
\fi
#2}}
\providecommand{\BIBdecl}{\relax}
\BIBdecl

\bibitem{nicolas2012brain}
L.~F. Nicolas~Alonso and J.~Gomez~Gil, ``Brain computer interfaces, a review,'' \emph{Sensors}, vol.~12, no.~2, pp. 1211--1279, 2012.

\bibitem{kannathal2005characterization}
N.~Kannathal, U.~R. Acharya, C.~M. Lim, and P.~Sadasivan, ``Characterization of {EEG}—a comparative study,'' \emph{Computer Methods and Programs in Biomedicine}, vol.~80, no.~1, pp. 17--23, 2005.

\bibitem{vourvopoulos2016usability}
A.~Vourvopoulos and S.~B.~I. Badia, ``Usability and cost-effectiveness in brain-computer interaction: is it user throughput or technology related?'' in \emph{Proc. of the 7th Augmented Human Int'l Conf. 2016}, 2016, pp. 1--8.

\bibitem{drwuTNSRE2016}
D.~Wu, V.~J. Lawhern, W.~D. Hairston, and B.~J. Lance, ``Switching {EEG} headsets made easy: {Reducing} offline calibration effort using active wighted adaptation regularization,'' \emph{{IEEE} Trans. on Neural Systems and Rehabilitation Engineering}, vol.~24, no.~11, pp. 1125--1137, 2016.

\bibitem{liu2025spatial}
D.~Liu, S.~Li, Z.~Wang, W.~Li, and D.~Wu, ``Spatial distillation based distribution alignment ({SDDA}) for cross-headset {EEG} classification,'' \emph{arXiv preprint arXiv:2503.05349}, 2025.

\bibitem{wan2023eegformer}
Z.~Wan, M.~Li, S.~Liu, J.~Huang, H.~Tan, and W.~Duan, ``{EEG}former: A transformer--based brain activity classification method using {EEG} signal,'' \emph{Frontiers in Neuroscience}, vol.~17, p. 1148855, 2023.

\bibitem{yang2023biot}
C.~Yang, M.~Westover, and J.~Sun, ``{BIOT}: Biosignal transformer for cross-data learning in the wild,'' \emph{Advances in Neural Information Processing Systems}, vol.~36, pp. 78\,240--78\,260, Dec. 2023.

\bibitem{kostas2021bendr}
D.~Kostas, S.~Aroca-Ouellette, and F.~Rudzicz, ``{BENDR}: Using transformers and a contrastive self-supervised learning task to learn from massive amounts of {EEG} data,'' \emph{Frontiers in Human Neuroscience}, vol.~15, p. 653659, 2021.

\bibitem{wang2024cbramod}
J.~Wang, S.~Zhao, Z.~Luo, Y.~Zhou, H.~Jiang, S.~Li, T.~Li, and G.~Pan, ``{CB}ra{M}od: A criss-cross brain foundation model for {EEG} decoding,'' \emph{arXiv preprint arXiv:2412.07236}, 2024.

\bibitem{jiang2024large}
W.-B. Jiang, L.-M. Zhao, and B.-L. Lu, ``Large brain model for learning generic representations with tremendous {EEG} data in {BCI},'' in \emph{Proc. Int'l Conf. on Learning Representations}, Vienna, Austria, May. 2024.

\bibitem{wang2025eegpt}
G.~Wang, W.~Liu, Y.~He, C.~Xu, L.~Ma, and H.~Li, ``{EEGPT}: Pretrained transformer for universal and reliable representation of {EEG} signals,'' Vancouver, Canada, Dec. 2025, pp. 39\,249--39\,280.

\bibitem{jiangneurolm}
W.~Jiang, Y.~Wang, B.-l. Lu, and D.~Li, ``Neurolm: A universal multi-task foundation model for bridging the gap between language and {EEG} signals,'' in \emph{The Thirteenth Int'l Conf. on Learning Representations}, Singapore, APR. 2025.

\bibitem{liu2025clean}
D.~Liu, Z.~Chen, and D.~Wu, ``{CLEAN-MI}: A scalable and efficient pipeline for constructing high-quality neurodata in motor imagery paradigm,'' \emph{arXiv preprint arXiv:2506.11830}, 2025.

\bibitem{li2024survey}
J.~Li, X.~Gu, S.~Qiu, X.~Zhou, A.~Cangelosi, C.~K. Loo, and X.~Liu, ``A survey of wearable lower extremity neurorehabilitation exoskeleton: Sensing, gait dynamics, and human--robot collaboration,'' \emph{IEEE Trans. on Systems, Man, and Cybernetics: Systems}, vol.~54, no.~6, pp. 3675--3693, 2024.

\bibitem{hermann2017paradigm}
B.~Hermann, D.~W. Loring, and S.~Wilson, ``Paradigm shifts in the neuropsychology of epilepsy,'' \emph{Journal of the Int'l Neuropsychological Society}, vol.~23, no. 9-10, pp. 791--805, 2017.

\bibitem{mokienko2014motor}
O.~Mokienko, L.~Chernikova, A.~Frolov, and P.~Bobrov, ``Motor imagery and its practical application,'' \emph{Neuroscience and Behavioral Physiology}, vol.~44, no.~5, pp. 483--489, 2014.

\bibitem{neuper2006motor}
C.~Neuper, G.~R. M{\"u}ller-Putz, R.~Scherer, and G.~Pfurtscheller, ``Motor imagery and {EEG}-based control of spelling devices and neuroprostheses,'' \emph{Progress in Brain Research}, vol. 159, pp. 393--409, 2006.

\bibitem{he2019transfer}
H.~He and D.~Wu, ``Transfer learning for brain-computer interfaces: A euclidean space data alignment approach,'' \emph{IEEE Trans. on Biomedical Engineering}, vol.~67, no.~2, pp. 399--410, 2019.

\bibitem{blankertz2007optimizing}
B.~Blankertz, R.~Tomioka, S.~Lemm, M.~Kawanabe, and K.-R. Muller, ``Optimizing spatial filters for robust {EEG} single-trial analysis,'' \emph{IEEE Signal processing magazine}, vol.~25, no.~1, pp. 41--56, 2007.

\bibitem{schirrmeister2017deep}
R.~T. Schirrmeister, J.~T. Springenberg, L.~D.~J. Fiederer, M.~Glasstetter, K.~Eggensperger, M.~Tangermann, F.~Hutter, W.~Burgard, and T.~Ball, ``Deep learning with convolutional neural networks for {EEG} decoding and visualization,'' \emph{Human Brain Mapping}, vol.~38, no.~11, pp. 5391--5420, 2017.

\bibitem{lawhern2018eegnet}
V.~J. Lawhern, A.~J. Solon, N.~R. Waytowich, S.~M. Gordon, C.~P. Hung, and B.~J. Lance, ``{EEGN}et: a compact convolutional neural network for {EEG}--based brain--computer interfaces,'' \emph{Journal of Neural Engineering}, vol.~15, no.~5, p. 056013, 2018.

\bibitem{mane2021fbcnet}
R.~Mane, E.~Chew, K.~Chua, K.~K. Ang, N.~Robinson, A.~P. Vinod, S.-W. Lee, and C.~Guan, ``{FBCN}et: A multi--view convolutional neural network for brain-computer interface,'' \emph{arXiv preprint arXiv:2104.01233}, 2021.

\bibitem{wang2023ifnet}
J.~Wang, L.~Yao, and Y.~Wang, ``{IFN}et: An interactive frequency convolutional neural network for enhancing motor imagery decoding from {EEG},'' \emph{IEEE Trans. on Neural Systems and Rehabilitation Engineering}, vol.~31, pp. 1900--1911, 2023.

\bibitem{tao2023adfcnn}
W.~Tao, Z.~Wang, C.~M. Wong, Z.~Jia, C.~Li, X.~Chen, C.~P. Chen, and F.~Wan, ``{ADFCNN}: attention-based dual-scale fusion convolutional neural network for motor imagery brain--computer interface,'' \emph{IEEE Trans. on Neural Systems and Rehabilitation Engineering}, vol.~32, pp. 154--165, 2023.

\bibitem{song2022eeg}
Y.~Song, Q.~Zheng, B.~Liu, and X.~Gao, ``{EEG} conformer: Convolutional transformer for {EEG} decoding and visualization,'' \emph{IEEE Trans. on Neural Systems and Rehabilitation Engineering}, vol.~31, pp. 710--719, 2022.

\bibitem{han2024edpnet}
C.~Han, C.~Liu, C.~Cai, J.~Wang, and D.~Qian, ``{EDPN}et: An efficient dual prototype network for motor imagery {EEG} decoding,'' \emph{arXiv e-prints}, pp. arXiv--2407, 2024.

\bibitem{steyrl2016random}
D.~Steyrl, R.~Scherer, J.~Faller, and G.~R. M{\"u}ller-Putz, ``Random forests in non--invasive sensorimotor rhythm brain-computer interfaces: a practical and convenient non-linear classifier,'' \emph{Biomedical Engineering/Biomedizinische Technik}, vol.~61, no.~1, pp. 77--86, 2016.

\bibitem{goldberger2000physiobank}
A.~L. Goldberger, L.~A. Amaral, L.~Glass, J.~M. Hausdorff, P.~C. Ivanov, R.~G. Mark, J.~E. Mietus, G.~B. Moody, C.-K. Peng, and H.~E. Stanley, ``Physiobank, physiotoolkit, and physionet: components of a new research resource for complex physiologic signals,'' \emph{Circulation}, vol. 101, no.~23, pp. e215--e220, 2000.

\bibitem{dreyer2023large}
P.~Dreyer, A.~Roc, L.~Pillette, S.~Rimbert, and F.~Lotte, ``A large {EEG} database with users’ profile information for motor imagery brain-computer interface research,'' \emph{Scientific Data}, vol.~10, no.~1, p. 580, 2023.

\bibitem{yi2014evaluation}
W.~Yi, S.~Qiu, K.~Wang, H.~Qi, L.~Zhang, P.~Zhou, F.~He, and D.~Ming, ``Evaluation of {EEG} oscillatory patterns and cognitive process during simple and compound limb motor imagery,'' \emph{PloS One}, vol.~9, no.~12, p. e114853, 2014.

\bibitem{Jayaram2018}
V.~Jayaram and A.~Barachant, ``{MOABB}: trustworthy algorithm benchmarking for {BCI}s,'' \emph{Journal of Neural Engineering}, vol.~15, no.~6, p. 066011, 2018.

\bibitem{zhou2016fully}
B.~Zhou, X.~Wu, Z.~Lv, L.~Zhang, and X.~Guo, ``A fully automated trial selection method for optimization of motor imagery based brain--computer interface,'' \emph{PloS One}, vol.~11, no.~9, p. e0162657, 2016.

\bibitem{lee2019eeg}
M.-H. Lee, O.-Y. Kwon, Y.-J. Kim, H.-K. Kim, Y.-E. Lee, J.~Williamson, S.~Fazli, and S.-W. Lee, ``{EEG} dataset and openbmi toolbox for three {BCI} paradigms: An investigation into {BCI} illiteracy,'' \emph{GigaScience}, vol.~8, no.~5, p. giz002, 2019.

\bibitem{cho2017eeg}
H.~Cho, M.~Ahn, S.~Ahn, M.~Kwon, and S.~C. Jun, ``{EEG} datasets for motor imagery brain--computer interface,'' \emph{GigaScience}, vol.~6, no.~7, p. gix034, 2017.

\bibitem{tangermann2012review}
M.~Tangermann, K.-R. M{\"u}ller, A.~Aertsen, N.~Birbaumer, C.~Braun, C.~Brunner, R.~Leeb, C.~Mehring, K.~J. Miller, G.~R. M{\"u}ller-Putz \emph{et~al.}, ``Review of the {BCI} competition {IV},'' \emph{Frontiers in neuroscience}, vol.~6, p.~55, 2012.

\bibitem{faller2012autocalibration}
J.~Faller, C.~Vidaurre, T.~Solis-Escalante, C.~Neuper, and R.~Scherer, ``Autocalibration and recurrent adaptation: Towards a plug and play online {ERD}-{BCI},'' \emph{IEEE Trans. on Neural Systems and Rehabilitation Engineering}, vol.~20, no.~3, pp. 313--319, 2012.

\bibitem{leeb2007brain}
R.~Leeb, F.~Lee, C.~Keinrath, R.~Scherer, H.~Bischof, and G.~Pfurtscheller, ``Brain--computer communication: motivation, aim, and impact of exploring a virtual apartment,'' \emph{IEEE Trans. on Neural Systems and Rehabilitation Engineering}, vol.~15, no.~4, pp. 473--482, 2007.

\bibitem{alexandre2006commande}
B.~Alexandre, ``Commande robuste d’un effecteur par une interface cerveau--machine {EEG} asynchrone,'' \emph{Universit{\'e} de Grenoble}, 2006.

\bibitem{maeder2012pre}
C.~L. Maeder, C.~Sannelli, S.~Haufe, and B.~Blankertz, ``Pre-stimulus sensorimotor rhythms influence brain--computer interface classification performance,'' \emph{IEEE Trans. on Neural Systems and Rehabilitation Engineering}, vol.~20, no.~5, pp. 653--662, 2012.

\bibitem{drwuMITLBCI2022}
D.~Wu, X.~Jiang, and R.~Peng, ``Transfer learning for motor imagery based brain-computer interfaces: A tutorial,'' \emph{Neural Networks}, vol. 153, pp. 235--253, 2022.

\bibitem{drwuPIEEE2023}
D.~Wu, B.-L. Lu, B.~Hu, and Z.~Zeng, ``Affective brain-computer interfaces ({aBCIs}): A tutorial,'' \emph{Proc. of the IEEE}, vol.~11, no.~10, pp. 1314--1332, 2023.

\bibitem{drwuEA2025}
D.~Wu, ``Revisiting {E}uclidean alignment for transfer learning in {EEG}-based brain-computer interfaces,'' \emph{Journal of Neural Engineering}, vol.~22, p. 031005, 2025.

\end{thebibliography}

\end{document}